\documentclass[journal]{IEEEtran}

\usepackage{cite}

\usepackage[dvipsnames]{xcolor}
\usepackage{amssymb}
\usepackage{pifont}
\newcommand{\cmark}{\color{Green}\ding{51}}
\newcommand{\xmark}{\color{red}\ding{55}}

\usepackage{adjustbox}
\usepackage{multirow}
\usepackage{amsmath}
\usepackage{bm}
\usepackage{subfig}
\usepackage{algorithm}
\usepackage{algorithmic}
\usepackage{graphicx}

\usepackage{url}

\begin{document}

\title{FREEDOM: Target Label \& Source Data \& Domain Information-Free Multi-Source Domain Adaptation \\ for Unsupervised Personalization}

\author{Eunju Yang,~\IEEEmembership{Student Member,~IEEE,}
        Gyusang Cho,
        and~Chan-Hyun Youn,~\IEEEmembership{Senior~Member,~IEEE}
\thanks{Eunju Yang, Gyusang Cho, and Chan-Hyun Youn are with the Department
of Electrical Engineering, KAIST, Korea,
e-mail: \{yejyang, gyusang.cho, chyoun\}@kaist.ac.kr. \\
Manuscript received Feb 10, 2023. This work has been submitted to the IEEE for possible publication. Copyright may be transferred without notice, after which this version may no longer be accessible.}}

\markboth{\tiny
    This work has been submitted to the IEEE for possible publication. Copyright may be transferred without notice, after which this version may no longer be accessible
}%
{Shell \MakeLowercase{\textit{et al.}}: Bare Demo of IEEEtran.cls for IEEE Journals}

\maketitle

\begin{abstract}
From a service perspective, Multi-Source Domain Adaptation (MSDA) is a promising scenario to adapt a deployed model to a client's dataset. It can provide adaptation without a target label and support the case where a source dataset is constructed from multiple domains. However, it is impractical, wherein its training heavily relies on prior domain information of the multi-source dataset --- how many domains exist and the domain label of each data sample. Moreover, MSDA requires both source and target datasets simultaneously (physically), causing storage limitations on the client device or data privacy issues by transferring client data to a server. For a more practical scenario of model adaptation from a service provider's point of view, we relax these constraints and present a novel problem scenario of \underline{T}hree-\underline{F}ree \underline{D}omain \underline{A}daptation, namely \textbf{TFDA}, where 1) target labels, 2) source dataset, and mostly 3) source domain information (domain labels + the number of domains) are unavailable. 
Under the problem scenario, we propose a practical adaptation framework called \textit{FREEDOM}. It leverages the power of the generative model, disentangling data into \textit{class} and \textit{style} aspects, where the style is defined as the class-independent information from the source data and designed with a nonparametric Bayesian approach. In the adaptation stage, FREEDOM aims to match the source class distribution with the target's under the philosophy that class distribution is consistent even if the style is different; after then, only part of the classification model is deployed as a personalized network. As a result, FREEDOM achieves state-of-the-art or comparable performance even without domain information, with reduced final model size on the target side, independent of the number of source domains.
\end{abstract}


\begin{figure}[t!]
    \centering
    \includegraphics[width=.86\linewidth]{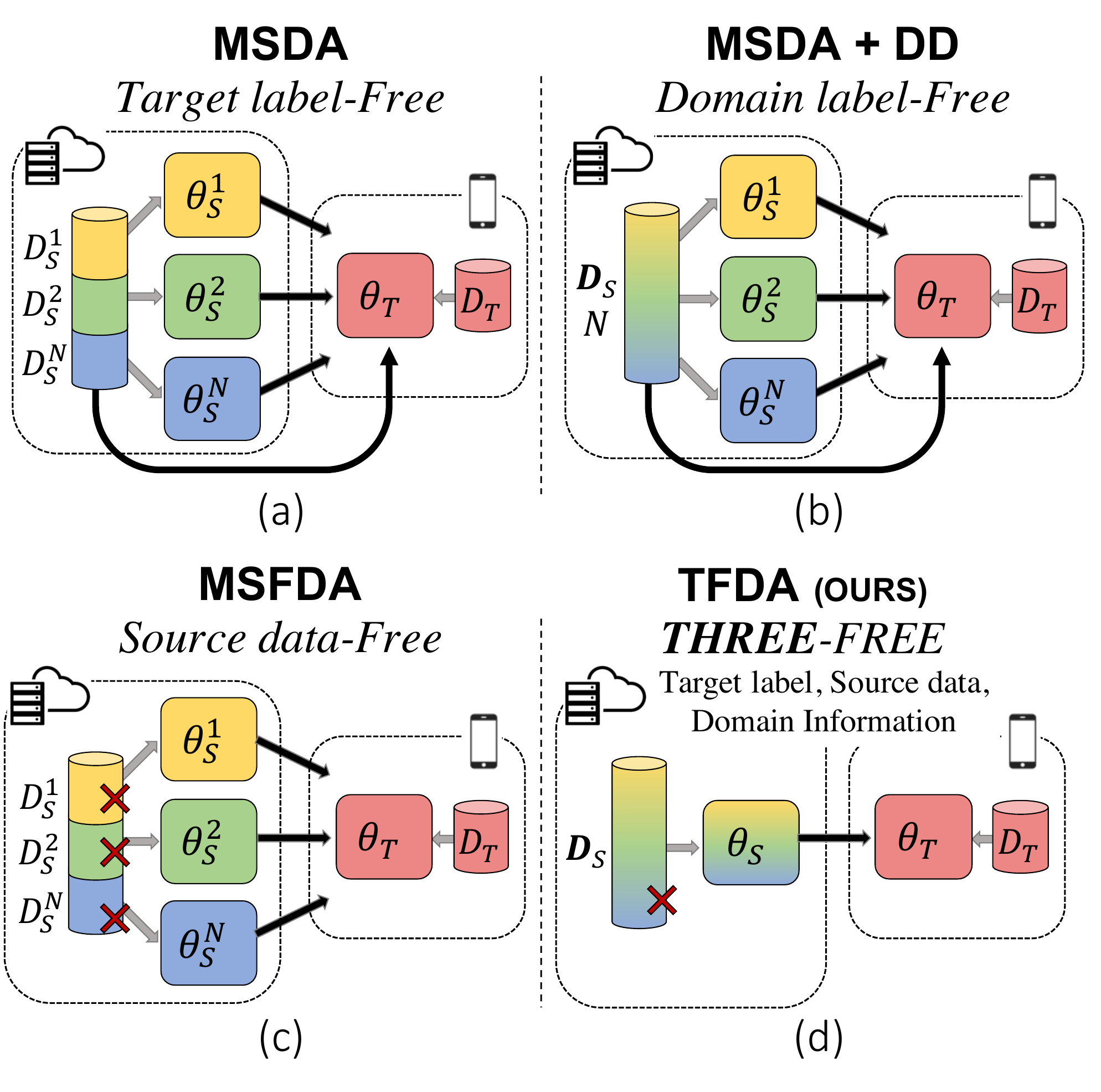}
    \caption{
        Comparison for several MSDA scenarios:  \textbf{(a) MSDA} trains target model with both multi-source and target dataset. \textbf{(b) MSDA+ Domain Discovery (DD)} conducts MSDA with source dataset without domain identifiers, while the number of source domains is a necessity. 
    \textbf{(c) MSFDA } exploits the source models in adaptation without a source dataset.
    \textbf{(d) TFDA (ours)} trains the target model only with the source model trained  without any information on how many domains and which domains are contained. 
    }
    \label{fig:scenario_comparison}
    \vspace{-5mm}
\end{figure}

\IEEEpeerreviewmaketitle

\section{Introduction}
\label{sec:introduction}
\IEEEPARstart{T}{he} domain shift problem caused by clients' dissimilar environments is one of the common obstacles for deep-learning-based service providers, as the applications are known to be data-dependent.  This problem originates from the distribution discrepancy between the client (\textit{target}) and server (\textit{source})-side datasets \cite{data_bias}. Additional adaptation with client data can be an alternative, but providing additional annotation to client data is burdensome in most cases. As a possible workaround, unsupervised domain adaptation (UDA) \cite{gretton2009covariate, ganin2015unsupervised}  and its downstream, multi-source domain adaptation (MSDA) \cite{DCTN_Xu_2018_CVPR, LtC_MSDA_wang2020learning} aim to adapt a model to an unlabeled target by leveraging labeled source dataset. 
Especially, MSDA considers the more plausible situation wherein it presumes the source dataset consists of samples from multiple domains.

Despite these technological advances, many factors still exist to consider when projecting real-world service scenarios onto MSDA's. Because of privacy issues on both source and target data, it is almost forbidden to transfer the dataset to each other. In other words, the client's unlabeled data can not be transferred to the server and vice versa. Moreover, sending multiple source datasets to the client may suffer storage limitations. Recent Source-Free UDA (SFUDA) has been introduced to address this situation by only sending a source-side model, not the dataset \cite{shot-v119-liang20a, usfda, yang2020bait, kim2021domain}. Multi-source-Free domain adaptation (MSFDA) approaches are also explored to support the multi-source cases \cite{ahmed2021decision, Caidar}. 

Existing MSFDA approaches \cite{Caidar, ahmed2021decision} usually train multiple models with each source dataset to weave them for the target, requiring domain information as prior knowledge. However, there are two additional factors to take into account: 1) maintaining domain labels is pricey; 2) handcrafted information on the number of domains in the training dataset can be overwhelming prior.
It is a well-known problem of domain adaptation that the domain information can be unprovided \cite{hoffman2012discovering, Carlucci2019Domain}. For example, a practitioner can collect training datasets from multiple channels \cite{mancini2018boosting}, in which the number of domains could be intractable. Besides, the number of domains can be overwhelming prior; a single dataset can consist of multiple latent domains. Treating the dataset that is believed to be a single domain as a single domain may not be optimal \cite{Xiong2014Latent, Li2020Discovering}.

In this paper, we relax the unhandled condition for domain information along with the MSFDA scenario, coined three-free domain adaptation (TFDA) --- domain adaptation scenario free from 1) target label, 2) source dataset at adaptation time, and 3) domain information, which is more pragmatic than previous scenarios described in Figure \ref{fig:scenario_comparison}. Here, domain information embraces both multi-source domain labels and the number of source domains. 
Under the scenario of the TFDA, we propose a three-\textbf{FREE} \textbf{DOM}ain adaptation method termed FREEDOM that trains a single model from a compound multi-source dataset and deploys it to a client, supporting unsupervised adaptation to the client dataset.
Since domain information is not provided and the target adaptation should be endowed without source datasets, we propose peripheral modules to transfer knowledge. We define `\textit{style}' as the remainder after subtracting typical class knowledge from the data; style is the knowledge that is the same as or includes the domain we usually believe. We train two encoders and a decoder to disentangle class and style embeddings from the given data while reconstructing its marginal distribution. As a remedy to handle domain information-free, we adopt nonparametric Bayesian as a prior for the style encoder. For the target adaptation, FREEDOM leverages the trained encoders and decoder from the source side and modulates the class encoder to transform a target input into the most likely embedding on the original class space while freezing the classifier layer.  
The ultimate goal of FREEDOM is to adapt the class encoder with hypothesis transfer  \cite{shot-v119-liang20a}. Thus, style encoders and decoders are exploited only to force stable adaption in a self-supervised manner and are eventually discarded after the tuning. Therefore, FREEDOM can have a lighter inference network than the MSFDAs, of which model size depends on the number of source domains \cite{ahmed2021decision,Caidar}. We summarize our contributions as follows:

\begin{itemize}
    \item We present a more pragmatic paradigm of Multi Source-Free Domain Adaptation with no domain information (domain labels + the number of domains), namely Three-Free Domain Adaptation (TFDA).
    \item We propose a disentangling-based FREEDOM with a novel alternating adaptation method to match the source and target class distribution; it exemplifies how to employ a generative model in source-free domain adaptation.
    \item The final adaptation model of FREEDOM's size is independent of the number of source domains, reducing the final personalized model without additional operation.
\end{itemize}

\begin{table}[t!]
\caption{Summary of the scenario comparisons.}
\begin{adjustbox}{width=\columnwidth,center}
\begin{tabular}{cccccc}
\hline
\multirow{3}{*}{} &
  \multirow{3}{*}{\begin{tabular}[c]{@{}c@{}}Multiple\\ Source\\ Domains\end{tabular}} &
  \multirow{3}{*}{\begin{tabular}[c]{@{}c@{}}\textbf{Target}\\ \textbf{Label} \\ \textbf{Free} \end{tabular}} &
  \multirow{3}{*}{\begin{tabular}[c]{@{}c@{}}\textbf{Source}\\ \textbf{Data} \\ \textbf{Free} \end{tabular}} &
  \multicolumn{2}{c}{\textbf{Domain Information} \textbf{Free}} \\ \cline{5-6} 
 &
  &
  &
  &
  \multirow{2}{*}{\begin{tabular}[c]{@{}c@{}}Domain \\ Label\end{tabular}} &
  \multirow{2}{*}{\begin{tabular}[c]{@{}c@{}}Number of\\ Domains \end{tabular}} \\
              &        &        &        &        &        \\ \hline
UDA           & \xmark & \cmark & \xmark & \xmark & \xmark \\
MSDA           & \cmark & \cmark & \xmark & \xmark & \xmark \\
MSDA+DD        & \cmark & \cmark & \xmark & \cmark & \xmark \\
SFUDA          & \xmark & \cmark & \cmark & \xmark & \xmark \\
MSFDA          & \cmark & \cmark & \cmark & \xmark & \xmark \\
TFDA (ours) & \cmark & \cmark & \cmark & \cmark & \cmark \\ \hline
\end{tabular}
\end{adjustbox}
\label{tab:scenario_comparison}
\end{table}

\begin{figure*}[t!]
    \centering
    \includegraphics[width=.95\linewidth]{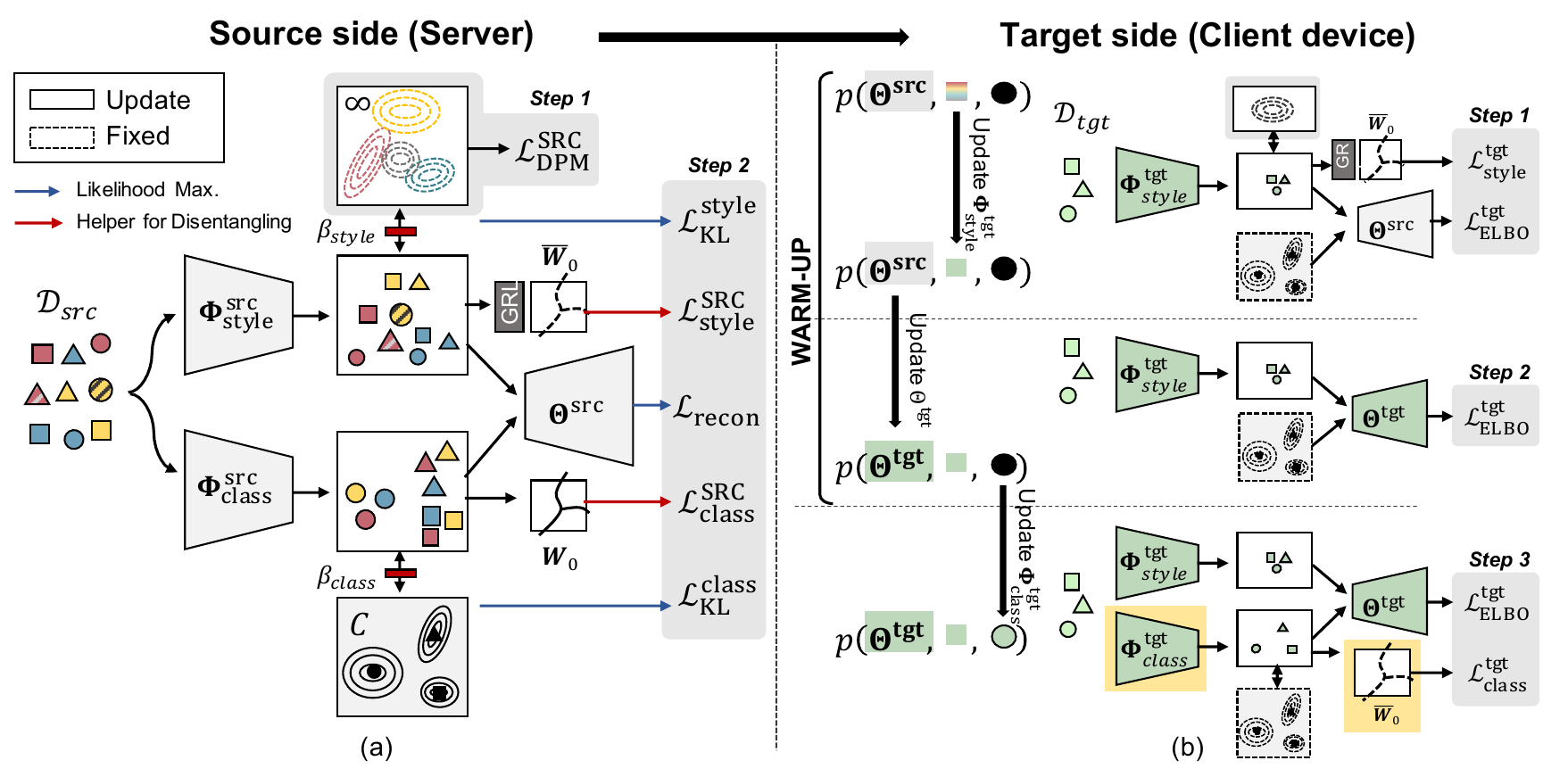}
    \caption{Overview of FREEDOM framework: (a) On the \textbf{source side}, FREEDOM trains disentangling networks using a multi-source dataset with two steps: finding prior distribution for style embeddings and finding the style and class embedding space while making them mutually independent and finding class prior distributions. The number of class embedding's prior is fixed as $C$, but the style embedding's prior cannot be configurable in advance but inferred throughout suboptimization from the given style embedding space. (b) On the \textbf{target side}, the style encoder,  decoder, and class encoder are trained in turn, except for the classifier and the class-conditional distribution. Every element with a dashed line is not involved in the adaptation (i.e., fixed). After the adaptation, layers in the yellow box are only deployed as a final personalized network.}
    \label{fig:freedom_overview}
    \vspace{-3mm}
\end{figure*}

\section{Related Works}
In this section, we introduce related works and provide comparisons across various MSDA scenarios in Table \ref{tab:scenario_comparison} to clarify the position of this study.

\textbf{Unsupervised Domain Adaptation}  aims to boost the accuracy of unlabeled targets by exploiting labeled source data. To this end, the datasets are used to learn features that can reduce the gap between domains represented by $\mathcal{H}$-divergence \cite{ben2010theory, mansour2009domain}. 
Two popular streams for minimizing the gap measure the discrepancy between the two domains \cite{DAN15, sun2017correlation} and using the adversarial training method \cite{ganin2015unsupervised}. The discrepancy-based method performs optimization by calculating a metric such as maximum mean discrepancy (MMD) \cite{gretton2012mmd}. Adversarial training employs a gradient reversal layer (GRL) to find a feature space that does not differentiate between domains and makes classification well \cite{ganin2015unsupervised, rangwani2022closer}. 
Furthermore, other generative model-based studies have been conducted for domain alignment \cite{cai2019disentangled, sankaranarayanan2018generate}. 
However, since they all presume a single source is given only, it is not practical in the real world.

\textbf{Multi-Source Domain Adaptation (MSDA)} handles unsupervised domain adaptation employing a source dataset with multiple domains, so it should consider domain discrepancies among various sources as well as domain gaps between the source and target. 
The main branch of the MSDA is the hypothesis combination, where each pair of a single source and the target is used in finding a hypothesis first, and the ultimate model for the target is implemented by their weighted mixture. Mansour et al. \cite{mansour2009domain} and Hoffman et al \cite{hoffman2018algorithms} presented the theoretical support of this hypothesis mixture for MSDA. Recent studies following this lineage take the form of training a model for each source-target pair and ensemble them; The algorithm focuses on how to find a common hypothesis for each pair and how to combine them.
For pair training, adversarial learning \cite{DCTN_Xu_2018_CVPR, MDDA_Zhao_Wang_Zhang_Gu_Li_Song_Xu_Hu_Chai_Keutzer_2020} or moment matching \cite{M3SDA_Peng_2019_ICCV} is widely adopted; For weight assignment, perplexity score \cite{DCTN_Xu_2018_CVPR}, weighted averaging \cite{M3SDA_Peng_2019_ICCV}, or Wasserstein distance \cite{MDDA_Zhao_Wang_Zhang_Gu_Li_Song_Xu_Hu_Chai_Keutzer_2020} is utilized. 

Unlike these, \cite{LtC_MSDA_wang2020learning} extracts prototypes from multiple sources as another form of knowledge.
Another branch is to train a single feature extractor across multiple domains. For example, \cite{venkat2020your} implicitly aligns all domain distributions by adopting multiple classifiers while sharing a feature extractor. \cite{mancini2018boosting} trains a network with mDA layers that can provide domain-wise normalization, generating a network with a normalization layer with a different moment for each domain.
These MSDA methods commonly require domain labels to make multi-domain to be aligned. However, identifying domains from a multi-domain dataset is pricey.

\textbf{Latent Domain Discovery (DD)} accounts for this practical issue of finding domain labels through \cite{hoffman2012discovering, gong2013reshaping, wu2019joint} or a discriminative network \cite{mancini2018boosting, mancini2019discovering}.
Hoffman et al. \cite{hoffman2012discovering} and  Wu et al. \cite{wu2019joint} adopt the Gaussian mixture model and hierarchical clustering to find domain identifiers. Meanwhile, \cite{mancini2018boosting, mancini2019discovering} employ an additional branch for domain discrimination, where the inference result is directly used in the MSDA network. Even though these domain discovery studies alleviate the cost of labeling in the domain aspect, they still require knowledge of the number of source domains as a prior, so they are not entirely free from domain information, unlike FREEDOM.

\textbf{Source-Free Domain Adaptation} is introduced to handle a challenging situation where existing DAs always require an enormous volume of the source dataset (even from multiple domains). For example, \cite{shot-v119-liang20a} resolves the problem via hypothesis transfer with self-supervised pseudo labeling; \cite{kim2021domain} use self-entropy for pseudo-label selection. In another way, \cite{Li_2020_CVPR} generates target-like data in order for model adaptation. 
However, these all presume the single source situation, in which performance is crushed with multiple source domains.

Recent \textbf{Multi-Source Free Domain Adaptation (MSFDA)} studies deal with this via confidence-anchor \cite{Caidar} or hypothesis transfer with optimization-based ensembling \cite{ahmed2021decision}. However, despite their outstanding contributions, they still rely on domain labels and their target model's size increase as the number of source domains increase. Thus, in this paper, FREEDOM considers a more plausible situation where domain information is not given, and the target model is independent of the increase of source domains.

\section{FREEDOM}
FREEDOM aims to resolve the TFDA scenario, where a model is trained with a multi-source dataset without domain information and deployed into a client device to support adaptation with the unlabeled target dataset.
Let $\mathcal{D}_{src} = \{(\bm{x}_n, y_n)\}_{n=1}^{N_s}$ and  $\mathcal{D}_{tgt} = \{ {\tilde{\bm{x}}}_n\}_{n=1}^{N_t}$ denote the multi-source and target datasets; their data distributions are different. The client's model is adapted to $\mathcal{D}_{tgt}$, leveraging the deployed model without any source data sample. So, server-side training is the only way to determine which knowledge to transfer from the multi-source dataset $\mathcal{D}_{src}$. 
Following the assumption of TFDA, the source dataset may consist of training samples from multiple domains while the information is not configurable, which complicates the problem since domain-wise model training is not allowed and requires additional manipulation. Besides, it is desirable to hand over the burden of adaptation to the server as much as possible since the target adaptation procedure is assumed to be performed on limited hardware. Therefore, FREEDOM consists of two training procedures: 1) source-side (server) training and 2) target-side (client device) adaptation,  as described in Fig. \ref{fig:freedom_overview}. It is discerned from the precedent MSFDA studies \cite{ahmed2021decision, Caidar}, presuming that multiple models are given by regular training.

\textbf{The source-side algorithm} is required to learn beneficial information for target adaptation, which should also be agnostic to the domain information. To accomplish this, FREEDOM takes three pillars of philosophy. \textbf{First}, we posit that every input data consists of \textit{class} and \textit{style} knowledge and build a disentangling model comprised of two encoders and a decoder. From the question `Is domain information necessary?', we find that the domain information is auxiliary in achieving the primary goal, and we chiefly need common class knowledge. Thus, if we have a way to draw the gist knowledge, the handcrafted domain labels are unnecessary. Based on this, we define \textit{style} as a non-class aspect, which means a residual obtained by subtracting class information from the data distribution. 
\textbf{Second}, we discover the prior distributions of each class, which are exploited as the blueprint for the target's class space. Thus, we posit that a class embedding follows the Gaussian Mixture Model (GMM); the source-side algorithm finds moments of each class's Gaussian distribution. Then by regularization transfer, we can guide the target's class encoder to find the space.
\textbf{Finally}, we define the prior style distribution with a nonparametric Bayesian method to make it serve without information on the number of source domains; the source-side algorithm regards the style aspect following Dirichlet Process Mixture (DPM).

\textbf{The target-side adaptation} adopts hypothesis transfer where the classifier is fixed \cite{shot-v119-liang20a}, so we only have to match the class embedding space with the original embedding space.
To this end, we exploit the generative model given by the server and pseudo-label. 
Upon the entropy maximization from the pseudo-label, it adapts the classification model to the target by maximizing the evidence of the target. The rationale for this comes from the distribution of the target can be described with the compound of the intrinsic class aspect obtained from the multi-source dataset and the target's style aspect. To make it find target distribution stably, FREEDOM proposes alternating adaptation relying on the class prior. Figure \ref{fig:freedom_overview} summarizes the overall behavior of FREEDOM following the TFDA scenario.

\subsection{Probabilistic Graphical of FREEDOM}

\subsubsection{Generative model}
Before introducing the algorithms' details, we delineate the underlying generative model that consists of the FREEDOM framework. We posit that input $\bm{x}_n \in \mathbb{R}^D$ is generated from class embedding $\bm{z}_n^{\text{class}} \in \mathbb{R}^{H_c}$ and style embedding $\bm{z}_n^{\text{style}} \in \mathbb{R}^{H_s}$, where each embedding follows GMM and DPM, respectively.
 The generative model of observation $\bm{x}_n$ follows the process :

\noindent\textbf{1. Choose latent class embedding $\bm{z}_n^{\text{class}}$}
\begin{itemize}
    \item $y_n \sim \text{Mult}(\bm{\pi}^{\text{class}})$, where $\bm{\pi}^{\text{class}} \in \Delta^{C-1}$
    \item $\bm{z}_n^{\text{class}} \vert y_n \sim \mathcal{N}(\bm{z}\vert \bm{\mu}_{y_n}^{\text{class}}, \bm{\Sigma}_{y_n}^{\text{class}})$
\end{itemize}
\textbf{2. Choose latent style embedding $\bm{z}_n^{\text{style}}$}
\begin{itemize}
    \item $\bm{\pi}^{style} \vert \gamma \sim \text{GEM}(\gamma)$
    \item $s_n \vert \bm{\pi}^{\text{style}} \sim \text{Mult}(\bm{\pi}^{style})$
    \item $\bm{\mu}^{\text{style}}_s \sim \mathcal{N}(\bm{\mu}\vert 0, \mathbb{I})$
    \item $\sigma_{s,h}^{\text{style}} \sim \text{Gamma}(1, 1)$
    \item $\bm{z}_n^{\text{style}} \vert s_n \sim \mathcal{N}(\bm{z}\vert \bm{\mu}_{s_n}^{\text{style}}, \bm{\Sigma}_{s_n}^{\text{style}}), $ where $\bm{\Sigma}_{s_n}^{\text{style}} = \bm{\sigma}^{\text{style}}_{{s_n}} \cdot \mathbb{I}$
\end{itemize}
\textbf{3. Choose a data point from the two embeddings}
\begin{itemize}
    \item $\bm{x} \sim \mathcal{N}(\bm{x}\vert \bm{\mu}_x, \bm{\Sigma}_x)$, where $[\bm{\mu}_x, \log\bm{\Sigma}_x] = f_{\bm{\Theta}}([\bm{z}^{\text{class}}:\bm{z}^{\text{style}}])$. Here, $\Theta$ is the decoder parameter,
\end{itemize}
where all notations are summarized in Table \ref{tab:notation_summary}.

\begin{figure}[t!]
    \centering
    \includegraphics[width=0.75\linewidth]{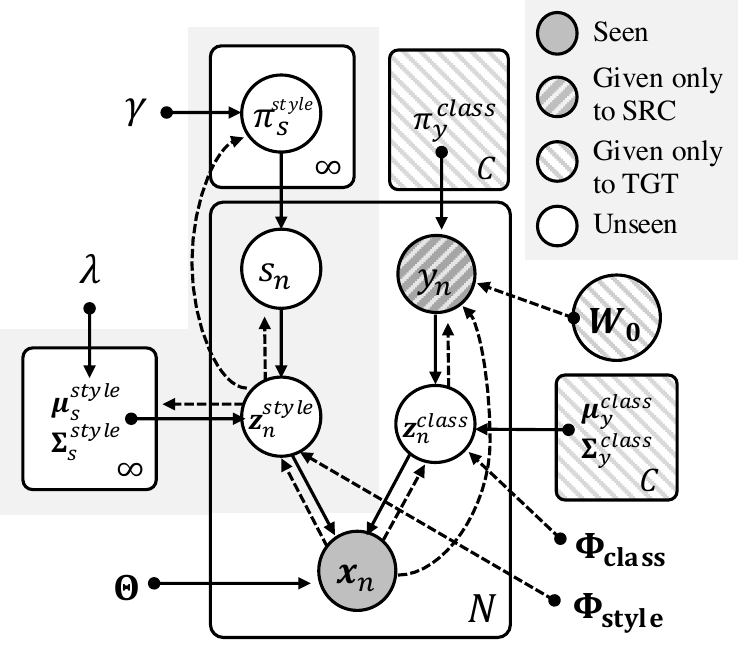}
    \caption{Probabilistic graphical model of FREEDOM. 
    The light gray box denotes the non-parametric Bayesian part, i.e., DPM, which is separately explored with given $\bm{z}_n^{\text{style}}$.}
    \label{fig:graphical_model}
\end{figure}

\begin{table}[t!]
\caption{Summary of notations in FREEDOM}
\begin{adjustbox}{width=\columnwidth,center}
\def\arraystretch{1.2}
\begin{tabular}{c|l}
\hline
\textbf{Notation}                  & \textbf{Description}                                                                                      \\ \hline
$\bm{x}_n$                         & Input image with dimension $D$                   \\
$\bm{z}_{n}^{\text{class}}$        & Class embedding from input $\bm{x}_n$, $\bm{z}_n^{\text{class}}\in\mathbb{R}^{H_c}$                       \\
$\bm{z}_{n}^{\text{style}}$        & Style embedding from input $\bm{x}_n$, $\bm{z}_n^{\text{style}}\in\mathbb{R}^{H_s}$                       \\
$H_c$                              & Dimension of class embedding                                                                              \\
$H_s$                              & Dimension of style embedding                                                                              \\
$y_n$                              & Class label of $\bm{x}_n$.                        \\
$\bm{\pi}^{\text{class}}$          & Prior distribution of the class labels                                                                    \\
$\bm{\mu}_{y_n}^{\text{class}}$    & Mean of $y_n$ class embedding distribution                             \\
$\bm{\Sigma}_{y_n}^{\text{class}}$ & Variance of the $y_n$ class embedding distribution                       \\
$\gamma$                           & hyperparameter for GEM                           \\
$\bm{\pi}^{\text{style}}$          & Prior distribution of the style identifier.                                     \\
$\bm{\mu}_{s_n}^{\text{style}}$    & Mean of style embedding distribution, specified with $s_n$                  \\
$\bm{\sigma}_{s_n}^{\text{style}}$ & $\bm{\sigma}_{s_n}^{\text{style}} = [\sigma_{s_n,h}]_{h=1}^{H_s}$                                         \\
$\bm{\mu}_x$                       & Mean of Gaussian distribution of $\bm{x}_n$                                                               \\
$\bm{\Sigma}_x$                    & Diagonal variance matrix of Gaussian distribution of $\bm{x}_n$                                           \\
$\bm{\Phi}_{\text{class}}$         & Class encoder parameter for $q(\bm{z}^{\text{class}}\vert\bm{x}_n)$         \\
$\bm{\Phi}_{\text{style}}$         & Style encoder parameter for $q(\bm{z}^{\text{class}}\vert\bm{x}_n)$         \\
$\bm{W}_0$                         & Classifier header of the inference model $g: \mathbb{R}^{\text{class}} \rightarrow \mathbb{R}^{C}$ \\ \hline
\end{tabular}
\end{adjustbox}
\label{tab:notation_summary}
\vspace{-3mm}
\end{table}

First, the class embedding, the hidden feature to discriminate into $C$ categories, follows a class-specific Gaussian distribution $\mathcal{N}(\bm{z}\vert \bm{y}_{y_n}^{\text{class}}, \bm{\Sigma}_{y_n}^{\text{class}})$ specified with its label $y_n \in [C]$. Specifically, the class label $y_n$ is determined by the multinomial distribution parameterized by $\bm{\pi}^{class} = \{\pi_y^{class}\}_{y=1}^{C} \in \mathbb{R}_+^C $, where $\sum_{y=1}^{C} \pi_y^{class} = 1$. Unlike class embeddings, which have explicit latent identifiers, it is challenging to know the number of mixtures for style embedding in advance. Thus, we postulate its prior distribution in a nonparametric Bayesian manner, especially DPM. As with the class embedding, let $\bm{\pi}^{style} = \{\pi^{style}_s\}_{s=1}^{\infty}$ be the prior probability of the style identifier, except having an infinite length; it is constructed with the Stick-Breaking process by additional random variable $\beta_s$, which follows the beta distribution. Then, we can define $\pi_s = \beta_s \prod_{l=1}^{s-1} (1-\beta_l)$; summing up the two processes, we can represent it with the Griffiths-Engen-McCloskey distribution (GEM). The given style identifier $s_n$ defines style-conditional distribution as Gaussian $\mathcal{N}(\bm{z}\vert\bm{y}_{s_n}^{\text{style}}, \bm{\Sigma}_{s_n}^{\text{style}})$, where its mean and variance follow Normal and Gamma distributions, respectively.
Finally, we can construct the data $\bm{x}_n$; we presume that the evidence follows Gaussian $\mathcal{N}(\bm{x} \vert \bm{\mu}_x, \bm{\Sigma}_x)$.
The parameters are derived by the decoder network $f_{\bm{\Theta}}$, i.e., the decoder returns the mean  $\bm{\mu}_x$ and variance $\bm{\Sigma}_x$ from the concatenated tensor of the two embeddings. Figure \ref{fig:graphical_model} describes the generative process; we can factorize the joint probability as follows:
\begin{equation}
\begin{aligned}
    &p(\bm{x}, \bm{z}^{class}, \bm{z}^{style}, y, s) = \\ 
    & p(\bm{x}\vert \bm{z}^{class}, \bm{z}^{style}) p(\bm{z}^{class}\vert y)p(y)
    p(\bm{z}^{style}\vert s)p(s).    
\end{aligned}
\label{eq:joint-prob}
\end{equation}

\subsubsection{Inference model}
We posit inference models of latent variables and find them throughout mean-field variational inference to discover the evidence distribution, where the joint variational distribution can be factorized as
\begin{equation}
\begin{aligned}
    &q(\bm{z}_n^{style}, \bm{z}_n^{class}, s_n, y_n\vert \bm{x}_n) \\
    &= q_{\bm{\Phi}^{style}}(\bm{z}_n^{style} \vert \bm{x}_n) q_{\bm{\Phi}^{class}}(\bm{z}_n^{class} \vert \bm{x}_n)
    q(s_n\vert\bm{x}_n) q(y_n\vert\bm{x}_n).
\end{aligned}
\label{eq:mean-field-assumption}
\end{equation}
First, for both class and style embedding, we presume the variational distributions, $q(\bm{z}^{\text{class}}\vert \bm{x})$ and $q(\bm{z}^{\text{style}}\vert \bm{x})$, follow the normal distribution like their generative models; the variational distributions' means and variances are inferred by encoder $\bm{\Phi}_{class}$ and $\bm{\Phi}_{style}$, respectively, i.e., $q(\bm{z}_n^{\text{class}} \vert \bm{x}_n) = \mathcal{N}(\bm{z};\hat{\bm{\mu}}^{\text{class}}, \hat{\Sigma}^{\text{class}})$ and  $[\hat{\bm{\mu}}^{\text{class}}, \log \hat{\bm{\Sigma}}^{\text{class}}] = f_{\Phi_{\text{class}}}(\bm{x})$. We also find inference models for other latent variables: style identifier $s_n$ and class label $y_n$; We propose that inference from inputs can be replaced with inferences from corresponding latent embeddings via Lemma 1, and we establish an inference model based on this. Inference on a style identifier and its style mode is assumed to be a DPM inference problem when style embedding is given. For class labels, it is replaced by an inference network $f_{\bm{W}_0}:\mathbb{R}^{H_c} \xrightarrow{\bm{W}_0} \mathbb{R}^{C}$ based on a supervised model. The final inference model for classification is a compound function of the class encoder and the classifier header, i.e. $f_{\bm{W}_0} \circ f_{{\bm{\Phi}}_{\text{class}}}(\bm{x}_n)$.
More details are provided in the subsequent section.

\subsection{Source-side Training}
We find all parameters for the generative and inference models on the source side as the way to knowledge transfer into a target.
Specifically, the training aims for two objectives: finding prior distribution on the class embedding space throughout evidence likelihood maximization and finding encoders to disentangle an input into style and class aspects.

\subsubsection{Evidence likelihood maximization}
The FREEDOM parameters are adjusted to maximize the log likelihood with the given multi-source domain samples $\mathcal{D}^{src}$. However, it is nontrivial to maximize it directly, for the term is intractable. As a workaround, we employ variational distribution $q(\bm{z}^{\text{style}}, \bm{z}^{\text{class}}, s, y\vert\bm{x})$ approximating the true distribution, and Jensen's inequality can substitute by the evidence lower bound (ELBO) maximization as follows:
\begin{align*}
    & \log p(\bm{x})\\
    & = \log \int \int \sum_{s} \sum_{y} p(\bm{x}, \bm{z}^{style}, \bm{z}^{class}, s, y)  d\bm{z}^{class} d\bm{z}^{style} \\
    & \geq 
    \mathbb{E}_{q}\Big[\log\frac{p(\bm{x}, \bm{z}^{style},\bm{z}^{class}, s, y)}{q(\bm{z}^{style},\bm{z}^{class},s,y\vert\bm{x})}\Big] = {\mathcal{L}_{\text{ELBO}}^{\text{SRC}} (\bm{x})}
\end{align*}
Then, by Eq. \ref{eq:joint-prob} and \ref{eq:mean-field-assumption}, we can factorize the source-side ELBO into three terms: 
\begin{equation}
\begin{aligned}
    \mathcal{L}_{\text{ELBO}}^{\text{SRC}} (\bm{x}, y) 
    & = \mathbb{E}_{q(\bm{z}^{\text{style}}, \bm{z}^{\text{class}}\vert\bm{x})} [\log p_{\bm{\Theta}}(\bm{x}\vert\bm{z}^{\text{style}}, \bm{z}^{\text{class}})] \\
    & \indent - \mathcal{D}_{\text{KL}}[q_{\bm{\Phi}_{\text{class}}}(\bm{z}^{\text{class}}, y\vert\bm{x})\vert\vert p(\bm{z}^{\text{class}}, y)] \\
    & \indent - \mathcal{D}_{\text{KL}}[q_{\bm{\Phi}_{\text{style}}}(\bm{z}^{\text{style}}, s\vert\bm{x})\vert\vert p(\bm{z}^{\text{style}}, s)] \\
    & := \mathcal{L}_{\text{recon}}(\bm{x}) -\mathcal{L}_{\text{KL}}^{\text{class}}(\bm{x},y) -\mathcal{L}_{\text{KL}}^{\text{style}}(\bm{x}),
\end{aligned}
\label{eq:freedom_elbo}
\end{equation}
where $\mathcal{D}_\text{KL}$ denotes the Kullback–Leibler (KL) divergence between the two distributions.
The first term represents the reconstruction loss ($\mathcal{L}_{\text{recon}}$); the remaining two imply the regularization term for class and style embedding to their respective prior, shorthand ($\mathcal{L}_{\text{KL}}^{\text{class}}$) and ($\mathcal{L}_{\text{KL}}^{\text{style}}$), respectively.

\textbf{The reconstruction loss} $\mathcal{L}_{\text{recon}}$ is computed by comparing the evidence and its reconstructed samples with the latent class and style embeddings taken from the two encoders. The latent embeddings are taken throughout the reparameterization trick \cite{kingma2013vae}, which fiddles with additional noise from the encoders' outputs, making the loss differentiable.

\textbf{The class regularizer}, the second term of Eq. \ref{eq:freedom_elbo}, can be further disassembled as 
\begin{equation}
\begin{aligned}
    & \mathcal{L}_{\text{KL}}^{\text{class}}(\bm{x}, y) := \mathcal{D}_{\text{KL}}[q(\bm{z}^{\text{class}}, y\vert\bm{x}) \vert\vert p(\bm{z}^{\text{class}}, y)] \\
    & = \mathbb{E}_q[\log p(\bm{z}^{\text{class}}\vert y)] + \mathbb{E}_q[\log p(y)] - \mathbb{E}_q[\log q(\bm{z}^{\text{class}}\vert \bm{x})] \\
    &  -\mathbb{E}_{q}[\log q(y\vert\bm{x})].
\end{aligned}
\label{eq:class_regularizer}
\end{equation}
\noindent Maximizing it enforces finding class-wise prior $p(\bm{z}^{class}\vert y)$ and the class encoder $f_{\bm{\Phi}^{class}}$, mapping an input to the class embedding space to satisfy the prior at once. 
We can streamline the loss function by exploiting the one-hot vector of the given class label $\bm{y} \in \mathbb{I}^{C}$ in place of 
the variational posterior of the class $q(y\vert \bm{x})$. The tractable form of class regularization loss is configurable in Appendix \ref{app:class_regularization_loss}.

\textbf{As the regularizer for the \textit{style} embedding}, however, its prior distribution is intractable due to the indefinite dimension, hindering finding the tractable form of the loss $\mathcal{L}_{\text{KL}}^{\text{style}}$. Specifically, the terms related to the style identifier $s$, e.g., $\mathbb{E}_{q}[\log p(\bm{z}^{\text{style}}\vert s)], \mathbb{E}_{q}[\log p(s)],$ and $ \mathbb{E}_{q}[\log q(s\vert\bm{x})]$. So, instead, we take a detour based on Lemma 1.

\underline{Lemma 1.} The optimal variational posterior of the style identifier $s$ is given as 
\begin{align*}
    q^*(s\vert\bm{x}) = \mathbb{E}_{q_{\bm{\Phi}^{\text{style}}}(\bm{z}^{\text{style}}\vert\bm{x})}[p(s\vert\bm{z}^{\text{style}})].
\end{align*}

The Lemma alludes that we can use the style embedding $\bm{z}^{\text{style}}$ from its variational posterior as a stepping stone to approximate the actual posterior of $s$. Inspired by this, we take an alternating update, decoupling the optimization into finding the style embedding's variational posterior and the prior distribution of the style embedding represented with the DPM. For the sake of explanation, let us impose the subscript $t$ to represent the optimization round. Then, instead of directly minimizing $\mathcal{L}^{\text{style}}_{\text{SRC}}$ concerning all hidden variables, we \textbf{(1)} explore the style distribution $p_{t}(\bm{z}^{\text{style}}) = \sum_{s} p(\bm{z}^{\text{style}}\vert s)p(s)$ using style embeddings from the variational posterior $q_{\bm{\Phi}^{\text{style}}_{t}}(\bm{z}^{\text{style}}\vert\bm{x})$ and \textbf{(2)} leverage it to update the style encoder, i.e., finding $q_{\bm{\Phi}_{t+1}^{\text{style}}}(\bm{z}^{\text{style}}\vert\bm{x})$.

\textit{(Step 1) Variational inference for style embedding's DPM}: 
Expressly, let $\bm{Z}_t^{\text{style}} = \{\bm{z}^{style}_n \vert \bm{z}^{\text{style}} \sim  q_{t}(\bm{z}^{\text{style}}\vert\bm{x}_n), \bm{x}_n \in \mathcal{D}^{\text{style}}\}$ and $\bm{\rho}_t = \{ \bm{\beta}_t, \bm{\theta}_t:=\{\bm{\mu_s^{\text{style}}}, \bm{\Sigma}_s^{\text{style}}\}, \bm{s}_t\}$ denote the set of style embeddings for the round $t$ and the set of hidden variables of the style embedding, respectively. Then, we find the posterior of $\bm{\rho}_t$ with $\bm{Z}_t^{\text{style}}$; since the distributions are still intractable and the massive evidence is given, we employ variational inference in finding DPM posterior throughout truncated stick-breaking approximation \cite{blei2006variational}.
The truncated stick-breaking distribution assumes that the total number of sticks representing $\bm{\beta}$ is fixed as $T$, which implies $q(\beta_T = 1) = 1$ and ${\pi}_s^{\text{style}} = 0$, $ \forall s > T$. Please note that this assumption is applied to variational distribution, not to the actual distribution; it alleviates the approximation difficulty. Then, the mean-field variational approximation for this DPM problem can be achieved by maximizing the following lower bound of DPM on $\bm{z}^{\text{style}}$,
\begin{equation}
\begin{aligned}
    & p(\bm{z}^{\text{style}}) \\
    & \geq  \mathbb{E}_{q}[\log p(\bm{z}^{\text{style}} \vert \bm{\mu}_{s}^{\text{style}}, \bm{\Sigma}_{s}^{\text{style}}, s)] 
    + \mathbb{E}_{q}[\log p(\bm{\mu}_s^{\text{style}})]  \\
    & +\sum_{n=1}^{N_s} \mathbb{E}_{q}[\log p(s_n \vert \bm{\beta})] + \mathbb{E}_{q}[\log p(\bm{\beta}\vert \gamma)]  - \mathbb{E}_{q}[\log q(\bm{\rho})] \\
    & := \mathcal{L}_{\text{ELBO}}^{\text{DPM}}(\bm{z}^{\text{style}})
\end{aligned}
\label{eq: dpm_elbo}
\end{equation}
where $q(\bm{\rho}) = \prod_{l=1}^{T-1} q_{\gamma_l}(\beta_l) \prod_{l=1}^{T} q_{\nu_{\mu_l}}(\mu_l) \prod_{h=1}^{H_s} q_{a_{lh}, b_{lh}}(\sigma_{lh}) \\ \prod_{n=1}^{N_s} q_{\phi_n}(s_n)$. Here, $q_{\gamma_l}$ is Beta distribution, $q_{\nu_{\mu_l}}$ is Normal distribution, $q_{a_{lh}, b_{lh}}(\sigma_{lh})$ is Gamma distribution, and $q_{\phi_n}(s_n)$ is multinomial distribution. Then, we find the optimal $\bm{\rho}^*_t$ maximizing $\mathcal{L}_{\text{ELBO}}^{\text{DPM}}$ throughout the coordinate ascent \cite{blei2006variational}.

\textit{(Step 2) Maximizing the style regularization term}: 
After finding the optimal $\bm{\rho}^*$ in \textit{(Step 1)},  it is exploited as an approximation of the prior distribution in calculating the style regularizer $\mathcal{L}_{\text{KL}}^{\text{style}}$, simplifying the problem with the finite dimension of the prior distribution.
Given the prior approximation, we should find only the variational parameter for the style embedding, that is, a style encoder $\bm{\Phi}_{\text{style}}$. Thus, we can remove irrelevant terms, simplifying the regularization term as
\begin{equation}
\begin{aligned}
    &\bar{\mathcal{L}}_{\text{KL}}^{\text{Style}}(\bm{x}, \bm{\beta}^*, \bm{\mu}^*, \bm{\Sigma}^*) \\
    & = \mathbb{E}_{q}[\log q(\bm{z}^{\text{style}}\vert \bm{x})] - \mathbb{E}_{q}[\log p(\bm{z}^{\text{style}}\vert \bm{\mu}^*, \bm{\Sigma}^*)].
\end{aligned}
\end{equation}

\begin{algorithm}[t!]
\caption{FREEDOM Training on \textbf{Source-side}}
\label{alg:source-side-training}
\textbf{Input}: Multi-source domain dataset $D_s$\\
\textbf{Parameter}: $C, \beta_{\text{low}}, \beta_{\text{high}}, l$\\
\textbf{Output}: $\theta_s$ 
\begin{algorithmic}[1] 
\STATE $ t \leftarrow 0$
\WHILE{not converge}
\STATE \textcolor{gray}{\underline{// \textbf{[STEP 1]} Finding DPM posterior}}
\STATE $\bm{Z}_t^{\text{style}} = \{\bm{z}_n^{\text{style}} \vert \bm{z}_n^{\text{style}} \sim  q_{\bm{\Phi}_t^{\text{style}}}(\bm{z}^{\text{style}}\vert\bm{x}_n), \bm{x}_n \in \mathcal{D}^{\text{style}} \}$
\STATE Find $\bm{\rho}^*_t $ maximizing $\mathcal{L}_{\text{ELBO}}^{\text{DPM}}(\bm{Z}_t^{\text{style}}) $ via coordinate ascent
\STATE $i \leftarrow 0$
\STATE \textcolor{gray}{\underline{// \textbf{[STEP 2]} Finding remaining parameters}}
\FOR{one epoch}
\STATE $(\beta_{\text{style}}, \beta_{\text{class}}) \leftarrow [(\beta_{\text{low}}, \beta_{\text{high}}), (\beta_{\text{low}}, \beta_{\text{low}})][i\%2]$
\STATE Compute $\nabla \mathcal{L}_{t}^{\text{src}}(\bm{x}, y, \beta_{\text{style}}, \beta_{\text{class}}, \bm{\rho}_t^*)$ (\ref{eq:src_loss})
\STATE Update $\Theta, \Phi_{\text{class}}, \Phi_{\text{style}}, \bm{W}_0, \{\bm{\mu}_y^{\text{class}}, \bm{\Sigma}_y^{\text{class}}\}_{y=1}^{C}, \bm{\pi}^{\text{class}}$ 
\STATE $\bar{\bm{W}}_0 \leftarrow \bm{W}_0$ ; $i \leftarrow i + 1$
\ENDFOR
\STATE $ t \leftarrow t + 1$
\ENDWHILE
\STATE \textbf{return} $\theta_s \leftarrow (\Theta, \Phi_{\text{class}}, \Phi_{\text{style}}, \{\bm{\mu}_y^{\text{class}}, \bm{\Sigma}_y^{\text{class}}\}_{y=1}^{C}, \bm{W}_0)$
\end{algorithmic}
\end{algorithm}

\subsubsection{Disentangling loss}
Besides the data likelihood maximization, FREEDOM achieves disentanglement from the original input without domain information. Therefore, the class and style embeddings should be independent while reconstructing the data. To this end, we control the hyperparameter of each regularizer in turn, inspired by \cite{jeong2019learning}. 
By first being strongly tied to the class embeddings' regularizer and being loosened later, we can control the route that the class encoder can take the information. 
To be more specific, the class encoder preferentially receives information from the class label. Then it obtains the rest of the information after the style encoder takes from the marginal distribution and vice versa.

In addition, we adopt two additional loss functions to clarify the knowledge independence between the class and style embedding.
The class helper is imposed to make the class encoder extract class-related knowledge, and we exploit the label smoothing for the loss function to calibrate the classifier $\bm{W}_0$. 
\begin{align*}
    \mathcal{L}_{\text{LS}}^{\text{class}} = - \sum_{n=1}^{N_s} \tilde{y} \cdot \log f_{\bm{W}_0}( f_{\bm{\Phi}_{\text{class}}}(\bm{x}_n)),
\end{align*}
where $\tilde{y} = y \cdot (1-l) + l/C$ with given calibration parameter $l$.
This calibration is conjugated later for the confidence-based filtering in target adaptation.
As a style helper, we take negative cross entropy by prepositioning a gradient reversal layer (GRL) \cite{ganin2015unsupervised} ahead of copied class hypothesis $W_0$. This helper loss affects the style encoder only, not the class hypothesis $W_0$. To this end, we use a trick to copy the hypothesis parameter $W_0$ as a style's header $\bar{W}_0$ without any update.
\begin{align*}
    \mathcal{L}_{\text{helper}}^{\text{style}} = - \sum_{n=1}^{N_s} y\cdot \log f_{\bar{\bm{W}}_0}(\mathcal{R}(f_{\Phi_{\text{style}}}(\bm{x}_n))),
\end{align*}
where $\mathcal{R}$ denotes the GRL layer. 

\subsubsection{Summary of Source-side training}
Summing all these up, the loss for the source-side training from the given multi-source dataset $\mathcal{D}_{\text{src}}$ and approximation of the style prior at round $t$ is summarized as follows:
\begin{equation}
\begin{aligned}
    & \mathcal{L}_{t}^{\text{src}}(\bm{x}, y, \beta_{\text{style}}, \beta_{\text{class}}, \bm{\rho}_t^*) \\
    & = - \mathbb{E}_{q_{\bm{\Phi}_t^{\text{style}}}(\bm{z}^{\text{style}}\vert \bm{x}) q_{\bm{\Phi}_t^{\text{class}}}(\bm{z}^{\text{class}}\vert\bm{x})} \Big[\log p_{\bm{\Theta}_t}(\bm{x}\vert\bm{z}^{\text{style}}, \bm{z}^{\text{class}}) \Big] \\
    & \;\;\; + \beta_{\text{style}} \cdot \bar{\mathcal{L}}_{\text{KL}}^{\text{style}}(\bm{x}, \bm{\beta}^*_t, \bm{\mu}^*_t, \bm{\Sigma}_t^*) \\
    & \;\;\; + \beta_{\text{class}} \cdot \mathcal{L}_{\text{KL}}^{\text{class}}(\bm{x},y) \\
    & \;\;\; +  \mathcal{L}_{\text{LS}}^{\text{class}} (\bm{x}, y, l ; \bm{W}_0) + \mathcal{L}_{\text{helper}}^{\text{style}} (\bm{x}, y; \bm{\bar{W}}_0).
\end{aligned}
\label{eq:src_loss}
\end{equation}
In summary, the source-side training of FREEDOM consists of two steps. First, it finds DPM parameters throughout truncated variational inference, and then, it minimizes $\mathcal{L}_{t}^{\text{src}}$ with two different weights on the style and class regularizer, in turn. Algorithm \ref{alg:source-side-training} delineates this procedure.

\subsection{Target-side Adaptation}
On the target side, it starts by taking FREEDOM's all parameters from the source side; it inherits most of the probabilistic model of the source, except that the class label is not observable, so we can leverage the loss functions defined in the previous section by tweaking them with pseudo labels. By the generative model of FREEDOM, we posit that the class-conditional distributions discovered from the source are reusable, while style embeddings should be substituted; if we can reuse the same class distribution with the same hypothesis, then the inference model's accuracy can be guaranteed. Thus, the main objective of target adaptation is to match the target class embedding's space with the source's one represented by the class prior distribution $\{\mathcal{N}(\bm{z};\bm{\mu}_y^{\text{class}}, \bm{\Sigma}_y^{\text{class}})\}_{y=1}^{C}$ --- let us call this `\textit{original}'. The class encoder should transform target data into the most likely embedding among the original space. To this end, FREEDOM uses 1) target likelihood maximization while sticking to the original and 2) sample selection by confidence and moment matching.

\subsubsection{Likelihood maximization with alternating adaptation}
We take advantage of the generative model in order to recover the class's original space.
Please remember that the class regularizer in Eq. (\ref{eq:class_regularizer}) forces the encoder to adhere to the prior distribution; by preserving the original prior for the class embedding, we can impose inertia to stabilize adaptation. 
In addition, if we can find an ideal target distribution consisting of the style knowledge of the target and class knowledge from the original, which can mimic the target distribution, then likelihood maximization is true of finding the class encoder mapping into the original space. 
It can guide the class encoders to avoid the pitfalls of non-original embedding spaces. We embody this by alternating updates of the FREEDOM's modules --- adapting style encoder  $\bm{\Phi}_{\text{style}}$, decoder $\bm{\Theta}$, and class encoder $\bm{\Phi}_{\text{class}}$ one by one.

\textbf{Target style encoder adaptation:} First, we find the style embedding of the target throughout style encoder adaptation. Specifically, we find style prior parameters $\bm{\rho}^*$ via maximizing Eq. (\ref{eq: dpm_elbo}) and employ the result in encoder adaptation throughout variants of Eq. (\ref{eq:src_loss}). The likelihood distribution is computed from the style embedding $\bm{z}^{\text{style}}$ drawn by the variational distribution $q_{\bm{\Phi}^{\text{style}}}(
\bm{z}\vert\bm{x})$ and class sample $\hat{\bm{z}}^{\text{class}}$ from the original space $\mathcal{N}(\bm{z};\bm{\mu}_{\hat{y}}^{\text{class}}, \bm{\Sigma}_{\hat{y}}^{\text{class}})$ based on its pseudo-label $\hat{y} = \arg\max f_{\bm{W}_0}(f_{\bm{\Phi}_{\text{class}}}(\bm{x}))$. The loss function of the style encoder adaptation is summarized as follows:
\begin{equation}
    \begin{aligned}
    \mathcal{L}^{\text{style}}_{\text{tgt}}(\tilde{\bm{x}}, \bm{\rho}^*) = & \mathcal{L}_{\text{helper}}^{\text{style}} (\tilde{\bm{x}}, \hat{y})
     + \bar{\mathcal{L}}_{\text{KL}}^{\text{style}}(\tilde{\bm{x}}, \bm{\beta}^*, \bm{\mu}^*, \bm{\Sigma}^*) \\
    & -\mathbb{E}_{q_{\bm{\Phi}_\text{style}}(\bm{z}^{\text{style}}\vert\bm{x})}[\log p_{\bm{\Theta}}(\bm{x} \vert \bm{z}^{\text{style}}, \hat{\bm{z}}^{\text{class}})] ,
    \end{aligned}
    \label{eq: target_style}
\end{equation}
where $\hat{\bm{z}}^{\text{class}} \sim \mathcal{N}(\bm{z};\bm{\mu}_{\hat{y}}^{\text{class}}, \bm{\Sigma}_{\hat{y}}^{\text{class}})$. Please note that the class embedding used in reconstruction loss is not drawn by the class encoder but by the original distribution. It tries to find the embedding, which is the remainder after subtracting the original distribution from the target.

\begin{algorithm}[t!]
\caption{FREEDOM Training on \textbf{Target-side}}
\label{alg:algorithm}
\textbf{Input}: Source-side parameters $\theta_s$, Target dataset $D_t$\\
\textbf{Output}: $\bm{\Phi}_{\text{class}}, \bm{W}_0$
\begin{algorithmic}[1] 
\STATE Initialize network parameters with $\theta_s$
\STATE Warm-up by repeating Steps 1 and 2
\WHILE{not converge}
\STATE{Filter out data, satisfying the conditions}
\STATE {Step 1-1.} Find $\bm{\rho}^*$ via coordinate ascent on $\mathcal{L}_{\text{ELBO}}^{\text{DPM}}$
\STATE {Step 1-2.} Update $\bm{\Phi}_{\text{style}}$ to minimize (\ref{eq: target_style})
\STATE {Step 2.} Update $\bm{\Theta}$ to minimize (\ref{eq:target_decoder})
\STATE {Step 3.} Update $\bm{\Phi}_{\text{class}}$ to minimize (\ref{eq:tgt_class})
\ENDWHILE
\STATE \textbf{return} $\theta_t \leftarrow (\bm{\Phi}_{\text{class}}, \bm{W}_0)$
\end{algorithmic}
\label{algo:target-side}
\end{algorithm}

\textbf{Target decoder adaptation:} After the style encoder adaptation, the decoder is tuned  to find an ideal target distribution consisting of the original class from the source and the target's style embedding.
The decoder only affects the reconstruction loss term in Eq. (\ref{eq:src_loss}), so its adaptation is conducted to minimize it; the reconstruction loss is computed similarly with the style encoder adaptation. In addition, we maximize the entropy from the reconstructed target $\hat{\bm{x}}$ in order to force it to take more credible information on the class space. Here is the loss function for the target decoder adaptation.
\begin{equation}
    \begin{aligned}
    \mathcal{L}^{\text{dec}}_{\text{tgt}} = & -\mathbb{E}_{q_{\bm{\Phi}_\text{style}}(\bm{z}^{\text{style}}\vert\bm{x})}[\log p_{\bm{\Theta}}(\bm{x} \vert \bm{z}^{\text{style}}, \hat{\bm{z}}^{\text{class}})] \\
    & - \sum_{n=1}^{N_t} \hat{y}\log f_{\bm{W}_0}\Big(f_{\bm{\Phi}_{\text{class}}}(\hat{\bm{x}})\Big),
    \end{aligned}
    \label{eq:target_decoder}
\end{equation}
where $\hat{\bm{x}} \sim p_{\bm{\Theta}}(\bm{x}\vert\bm{z}^{\text{style}}, \hat{\bm{z}}^{\text{class}})$.

\textbf{Target class encoder adaptation:}
So far, the style encoder and decoder have been updated to represent the target distribution implying the original space, so likelihood maximization intrinsically leads the class embedding space to the original. The target encoder is updated to minimize the following loss function:
\begin{equation}
    \begin{aligned}
        & \mathcal{L}_{\text{tgt}}^{\text{class}} (\tilde{\bm{x}}) \\
        & =  - \alpha^{\text{class}}_{\text{recon}} \cdot \mathbb{E}_{q(\bm{z}^{\text{style}}, \bm{z}^{\text{class}}\vert \tilde{\bm{x}})} [\log p(\tilde{\bm{x}}\vert\bm{z}^{\text{style}}, \bm{z}^{\text{class}})] \\ 
        & +  \alpha^{\text{class}}_{\text{KL}} \cdot \mathcal{L}_{\text{KL}}^{\text{class}}(\tilde{\bm{x}}, \hat{y}) 
        - \alpha^{\text{class}}_{\text{helper}} \cdot \sum_{n=1}^{N_t} \hat{y}\log f_{\bm{w}_0}\Big(f_{\bm{\Phi}_{\text{class}}}(\tilde{\bm{x}})\Big) 
    \end{aligned}
    \label{eq:tgt_class}
\end{equation}
The class regularization term of the ELBO loss is computed to force the class embedding to be tied to a class conditional prior, chosen by the pseudo-label. The entropy maximization adapts the class encoder to contain more information on its inference result.

\textbf{Alternating update for the target adaptation:} 
All of these adaptations are alternatingly conducted. One may think of tuning with the source-like optimization on Eq. (\ref{eq:src_loss}) using a pseudo label. However, the coercive optimization may find another class embedding space, which can maximize likelihood but does not accord with the original space, lowering the accuracy under the fixed hypothesis.   On the other hand, this sophisticated alternating optimization can narrow down the optimization objective. In order to enhance this confinement, we additionally adopt a warm-up step repeating style and decoder update prior to class encoder adaptation.
This warm-up provides the decoder's distribution to be more aligned with the target's, which clarifies the guide role of the reconstruction loss in encoder loss (\ref{eq:tgt_class}). That is to say; it ensures that the likelihood maximization does not fall into another class embedding space when updating the class encoder but toward the original.
The overall target side training is summarized in Algorithm \ref{algo:target-side}.

\subsubsection{Confident-based data selection}
The alternating adaptation algorithm heavily relies on the quality of the pseudo-label. Especially, at the beginning of the adaptation, the noise in the pseudo label is fatal, so we filter out confidential samples that are likely to be correct. To this end, FREEDOM exploits two different pieces of information. One is inference on the class based on the original, i.e., $\gamma^*_y = q^*_{\bm{\Phi_{\text{class}}}}(y\vert\bm{x})$, and the other is the confidence level of inference result drawn from the inference network $\hat{\bm{y}} = \text{SoftMax}(f_{\bm{W}_0}(f_{\bm{\Phi}_{\text{class}}}(\bm{x})))$. 

First, we check whether the inference results using the original and classifier are matched, i.e.,  $\arg\max{\gamma^*_y} = \arg\max \hat{\bm{y}}$; we only use the matching sample in adaptation. Second, we exploit the target sample where its confidence on the pseudo label is greater than the given confidence level $L$, i.e., $\max \hat{\bm{y}} \geq L$. In the evaluation section, we are going to validate the convergence of the confidence batch ratio across the target training. Here, the class label inference with the orignal $\gamma^*_y$ is computed with $\mathbb{E}_{q(\bm{z}^{\text{class}}\vert \bm{x})}[p(y\vert\bm{x})]$, 
where its detail and tractable form are described in Appendix \ref{app:gamma}.

\section{Experimental Evaluation and Discussion}
In this section, we validate FREEDOM with extensive experiments, from quantitative to qualitative analysis. Prior to describing the empirical results, let us expound on the general experiment settings.
After then, we introduce empirical analysis.

\subsection{Experiment Configuration}
\subsubsection{Dataset}
We evaluated FREEDOM with four popular MSFDA benchmarks: \texttt{Five-digit}, \texttt{Office}, \texttt{Office-Caltech}, and \texttt{Office-Home} datasets. The \texttt{Five-digit} dataset is a number-classification dataset with ten classes consisting of five domains, including MNISTM (\texttt{MM}), MNIST (\texttt{MT}), SVHN (\texttt{SV}), USPS (\texttt{UP}), and SYNNUM (\texttt{SYN}). The \texttt{Office} dataset 
\cite{saenko2010adapting} 
is a multi-domain classification dataset having 31 classes, which includes Amazon (A), DSLR (D), and Webcam (W) as domains. 
The \texttt{Office-Caltech} dataset is the intersection of the Office and Caltech datasets, consisting of 10 shared classes; its number of domains is four, including Caltech (C).
The \texttt{Office-Home} dataset \cite{Venkateswara_2017_CVPR} is another MSDA benchmark with 65 classes containing four domains, Art (A), Clipart (C), Product (P), and Real-World (R).

\subsubsection{Competing methods}
We compared FREEDOM with diverse variants of MSDA methods.
As baseline MSDA, we took MDAN \cite{MDAN_NEURIPS2018_717d8b3d}, DCTN\cite{DCTN_Xu_2018_CVPR}, M3SDA \cite{M3SDA_Peng_2019_ICCV}, MDDA \cite{MDDA_Zhao_Wang_Zhang_Gu_Li_Song_Xu_Hu_Chai_Keutzer_2020}, LtC-MSDA \cite{LtC_MSDA_wang2020learning}, STEM \cite{Nguyen_2021_ICCV}. They wholly focus on reducing the gap between multiple source domains and target datasets without any constraints on source-free or domain information-free.
Otherwise, mDA \cite{mancini2018boosting} and MEC \cite{mancini2019discovering} consider the case where domain labels are not given, but the number of source domains is given, which is pseudo-domain information free.

As a challenging objective, SFUDA approaches --- BAIT \cite{yang2020bait}, PrDA \cite{kim2021domain}, SHOT \cite{shot-v119-liang20a}, MA \cite{Li_2020_CVPR} ---  are also adopted as competing methods. For a fair comparison, the softmax average value on outputs from each source domain reported by \cite{Caidar} is demonstrated. Finally, MSFDA, which is most comparable with the TFDA scenario, is adopted as baselines, e.g., DECISION \cite{ahmed2021decision} and CAiDA \cite{Caidar}.

\def\arraystretch{1.2}
\begin{table}[t!]
\caption{Network architecture of FREEDOM for Five-Digit, Office, and Office-Home dataset.}
\begin{adjustbox}{width=\columnwidth,center}
\centering
\begin{tabular}{ccc}
\hline
           & \textbf{Five-Digit}                                                                                                                                                                                                                                                                                               & \textbf{Office / Office-Caltech / Office-Home}                                                                          \\ \hline 
\textbf{Backbone}   & -                                                                                                                                                                                                                                                                                                        & ResNet-50                                                                                     \\ \hline 
\textbf{Encoder}    & \begin{tabular}[c]{@{}c@{}}64 conv. 3x3, LeakyReLU\\ 64 conv. 3x3, LeakyReLU\\ MaxPool 2d 2x2\\ 64 conv. 3x3, LeakyReLU\\ 64 conv. 3x3, LeakyReLU\\ 64 conv. 3x3, LeakyReLU\\ MaxPool 2d 2x2\\ fc\_mu, 64x3x3, 512, Tanh\\ fc\_logvar, 64x3x3, 512, Tanh\end{tabular} & \begin{tabular}[c]{@{}c@{}}    fc 1024, 1024, ReLU   \\ fc\_mu 1024, 2000, Tanh\\ fc\_logvar 1024, 2000, Tanh\end{tabular} \\ \hline
\textbf{Decoder}    & \begin{tabular}[c]{@{}c@{}}fc, 1025, 64x6x6, Tanh\\ ConvTrans.2d 64x64x6x2, Tanh\\ ConvTrans.2d 64x64x6x1, Tanh\\ ConvTrans.2d 64x64x6x1, Tanh\\ Dropout p=0.5\\ ConvTrans.2d 64x64x3x1, Tanh\\ ConvTrans.2d 64x64x3x1, Tanh\\ ConvTrans.2d 64x3x3x1, Tanh\end{tabular}                & \begin{tabular}[c]{@{}c@{}} fc1 4000, 1024, ReLU \\ fc2 1024, 1024, ReLU   \end{tabular}             \\ \hline
\textbf{Classifier} & \begin{tabular}[c]{@{}c@{}}64 conv, 3x3, LeakyReLU\\ 64 conv, 3x3, LeakyReLU\\ 64 conv, 3x1, LeakyReLU\\ AvgPool 2d\\ fc 64, 10\end{tabular}                                                                                                                                                             & \begin{tabular}[c]{@{}c@{}}fc 2000, C\\ C = 31 or 10 or 65\end{tabular}                             \\ \hline
\end{tabular}
\end{adjustbox}
\label{tab:network}
\end{table}

\subsubsection{Implementation}
We implemented FREEDOM with PyTorch \cite{pytorch} and Scikit-learn \cite{scikit-learn}. In particular, we exploited the \texttt{BayesianGaussianMixture} module of Scikit-learn to construct DPM's variational inference (\texttt{line 5} in Algorithm \ref{alg:source-side-training} and \ref{algo:target-side}). 
When it comes to network architecture, we followed precedents. 
We adopted the network architecture from \cite{Nguyen_2021_ICCV} for the five-digit dataset while renovating it a little into FREEDOM's format ---  encoders and decoder structure.

We constructed the same network architecture using pre-trained ResNet as their backbone for the \texttt{Office}, \texttt{Office-Caltech}, and \texttt{Office-Home} datasets. We used the same structure for both style and class encoder; the input size of the decoder is twice that of each embedding, for the concatenation of the two is fed into the decoder. FC layers are used as encoders and decoders, where each encoder takes an embedding from the pre-trained ResNet-50. For the pre-trained parameter, we exploited  \texttt{ResNet50\_Weights.IMAGENET1K\_V2}, which is officially deployed in PyTorch. Since the backbone network is only used for input generation, it is not updated through the training, but encoders, the decoder, and the classifier layer are updated.
All network details are described in Table \ref{tab:network}.

We used Adam optimizer with $\beta_1, \beta_2 = 0.5, 0.99$  and StepLR scheduler for all training with a decay rate of 0.9.
We commonly adopt 0.1 and 5 as $\beta_{\text{low}}$ and $\beta_{\text{high}}$ for alternating training parameters. 
In the source-side training, we followed the pre-training strategy, widely adopted in variational model training \cite{jiang2017variational}, to train the network without any variational loss before starting the regular training. The label smoothing parameter $l$ is set to 0.15. More detailed hyper-parameters for each dataset are described in the following subsections.

\subsection{Evaluation on Five-digit dataset}

\subsubsection{Quantitative Analysis}
We foremost analyzed FREEDOM's performance on the \texttt{Five-digit} dataset. First, we train the source-side model with the rest of the domains except for the target, following the TFDA scenario. Then, for the source-side training, we trained the model 200 epochs based on Algorithm \ref{alg:source-side-training} with ten epochs of pre-training. After the source-side learning, the final model is deployed so as to adapt to the target. In target adaptation, we set the confidence level as 0.8; according to the confidence batch ratio, we imposed different weights on the class adaptation loss. For example, we imposed more weights on the class regularization term when the baseline model retrieves enough confident samples (\textbf{conf1} in Table \ref{tab:five-digit-exp-conf}); if not, we gave more weight to reconstruction loss (\textbf{conf2}).
Table \ref{tab:fivedigit}  summarizes the results. It contains FREEDOM's adaptation accuracy for each target domain and baseline, that is, test accuracy right after source-side training without any adaptation. For better comparison, its first three columns explain the characteristics of each method, which denote whether it supports multi-source (MS), source-free (SF), and domain information-free (DIF), respectively. Finally, all numerical results in the table describe the average value measured with four different random seeds, considering the characteristics of the variational model.

\begin{figure}[t!]%
    \centering
    \subfloat{{\includegraphics[width=.49\linewidth]{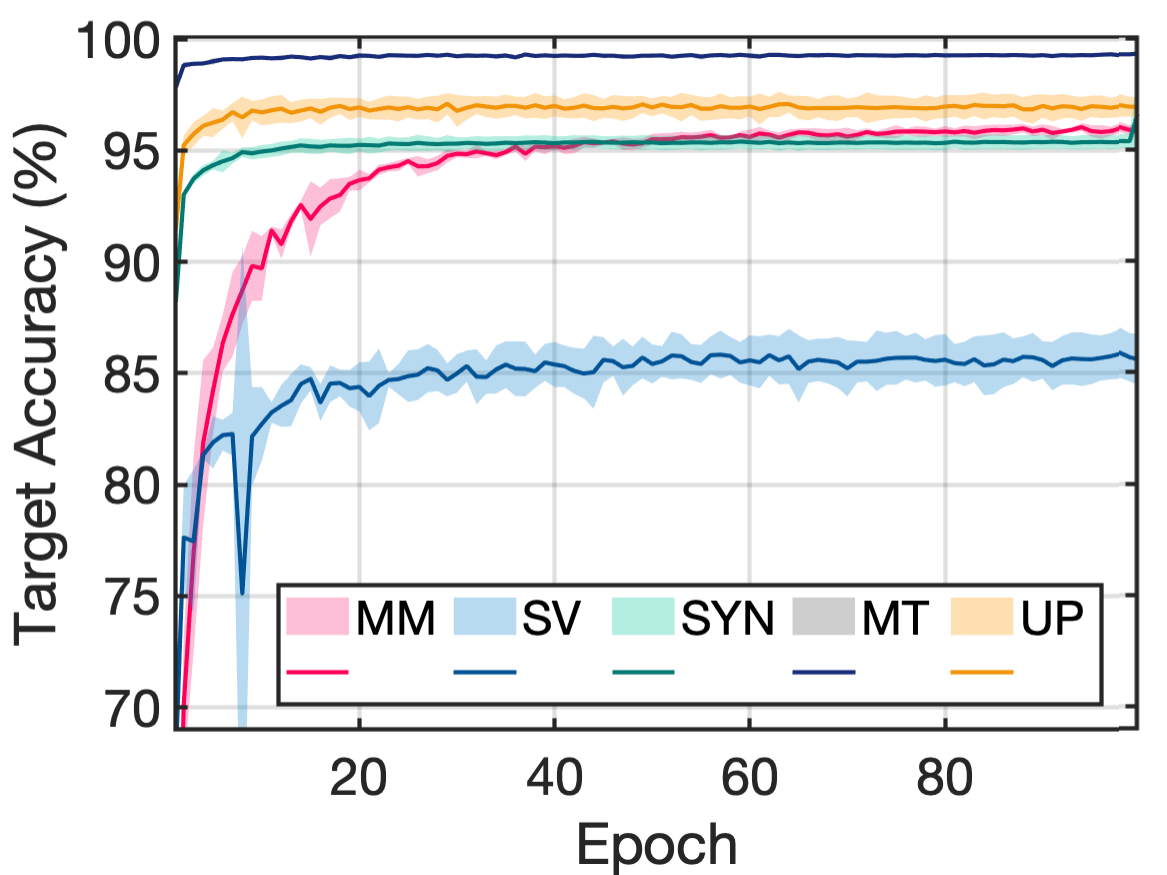} }} 
    \subfloat{{\includegraphics[width=.49\linewidth]{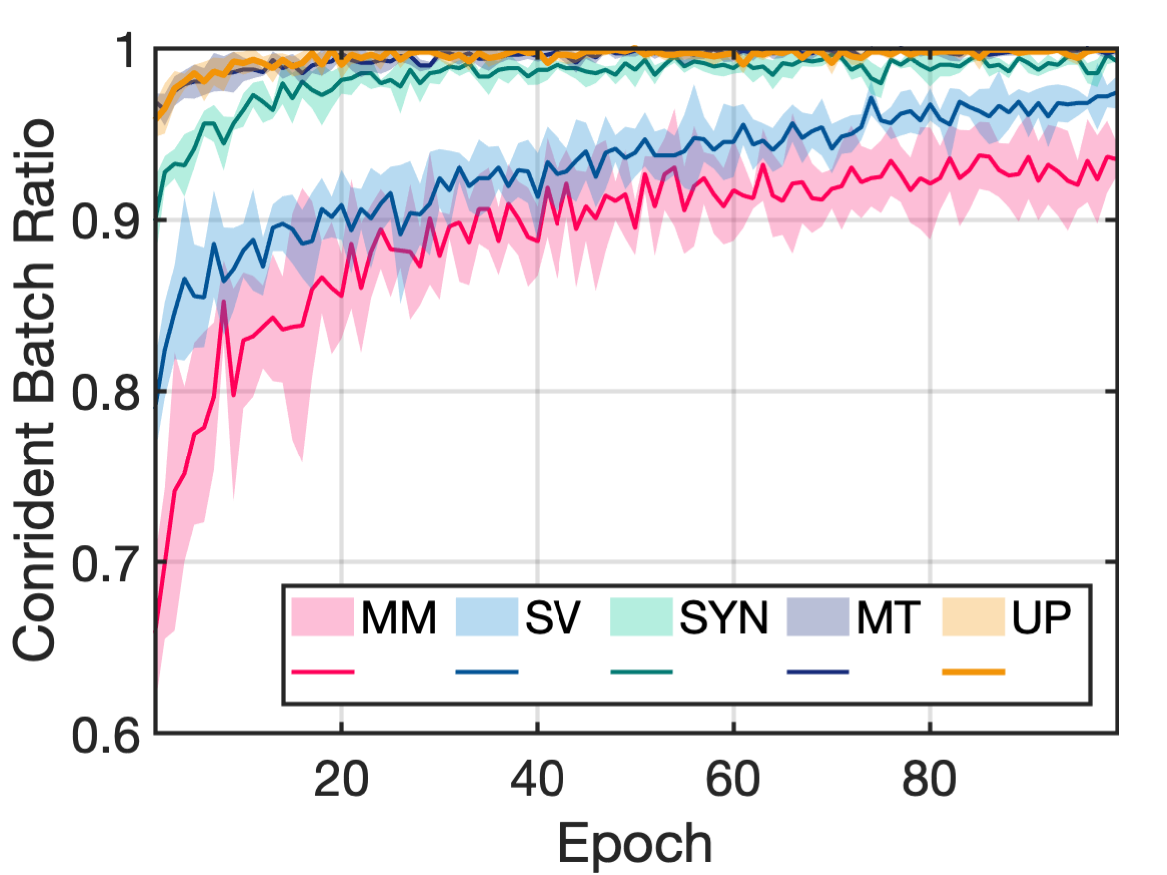} }}%
    \caption{Convergence in adaptation (\textbf{Left}) target accuracy convergence and (\textbf{Right}) confidence ratio convergence.}%
    \label{fig:fivedigit_convergence}%
\end{figure}

The results show that FREEDOM has, on average, the best performance for all target data, even though it satisfies the tighter constraints. The baseline performance is poor without adaptation because the distribution deviation between domains is significant, yet FREEDOM successfully adapts the model without target labels and source datasets. Fig. \ref{fig:fivedigit_convergence} demonstrates the convergence graph and the trend of the confidence batch ratio across the adaptation. The confident batch ratio is the number of confident samples normalized with its mini-batch size. Interestingly, the confidence batch ratio reflects its convergence, which can be used as a metric for unsupervised adaptation. The measure gives clues as to when to stop adaptation and hyperparameter tuning. 
For the case where the initial confidence ratio is low, e.g., \texttt{MM} and \texttt{SV}, it is desirable to give more weight to reconstruction loss rather than the regularization term in (\ref{eq:tgt_class}). Thus, by the metric, we applied \textbf{conf1} to \texttt{MT}, \texttt{UP}, and \texttt{SYN} targets and \textbf{conf2} to \texttt{MM} and \texttt{SV} targets, leading to outstanding performance in adaptation.

\def\arraystretch{1.2}
\begin{table}[b!]
\centering
\caption{Summary of experiment setting on \texttt{Five-digit} }
\begin{adjustbox}{width=.85\linewidth}
\begin{tabular}{ccccc}
\hline
\multicolumn{5}{c}{\textbf{Five-digit training general setting}}                                                                                                                                                                               \\ 
               & \textbf{Learning Rate}                                                      & \textbf{Epoch}                                                           & \textbf{Warmup}                                                              & \textbf{batch size}       \\ \hline 
{\textbf{SRC}}   & 1e-4                                                               & 200                                                             & 10                                                                  & 256              \\
{\textbf{TGT}}   & 5e-4                                                               & 100                                                             & 10                                                                  & 256              \\ \hline 
\multicolumn{5}{c}{\textbf{Five-digit target configuration}}                                                                                                                                                                                   \\ 
               & \begin{tabular}[c]{@{}c@{}}$\alpha^{\text{class}}_{\text{recon}}$\end{tabular} & \begin{tabular}[c]{@{}c@{}}$\alpha^{\text{class}}_{\text{KL}}$\end{tabular} & \begin{tabular}[c]{@{}c@{}}$\alpha^{\text{class}}_{\text{helper}}$\end{tabular} & confidence level \\ \hline
{\textbf{conf1}} & 1                                                                  & 5                                                               & 5                                                                   & 0.8              \\ 
{\textbf{conf2}} & 5                                                                  & 1                                                               & 5                                                                   & 0.8              \\\hline
\end{tabular}
\end{adjustbox}
\label{tab:five-digit-exp-conf}
\end{table}

\def\arraystretch{1.1}
\begin{table}[t!]
\caption{Evaluation results on \texttt{Five-digit} dataset. }
\begin{adjustbox}{width=.95\columnwidth,center}
\begin{tabular}{cccccccccc}
\hline
\textbf{Methods}        & \textbf{MS}           & \textbf{SF}           & \textbf{DIF}          & \textbf{\begin{tabular}[c]{@{}c@{}}$\rightarrow$ \\ MM\end{tabular}} & \textbf{\begin{tabular}[c]{@{}c@{}}$\rightarrow$ \\ MT\end{tabular}} & \textbf{\begin{tabular}[c]{@{}c@{}}$\rightarrow$ \\ SV\end{tabular}} & \textbf{\begin{tabular}[c]{@{}c@{}}$\rightarrow$ \\ UP\end{tabular}} & \textbf{\begin{tabular}[c]{@{}c@{}}$\rightarrow$ \\ SYN\end{tabular}}  & \textbf{Avg.}\\ \hline
MDAN             & \cmark & \xmark & \xmark & 69.5                                                                 & 98                                                                   & 69.2                                                                 & 92.4                                                                 & 87.4          & 83.3                                                        \\
DCTN             & \cmark & \xmark & \xmark & 70.5                                                                 & 96.2                                                                 & 77.6                                                                 & 92.8                                                                 & 86.8          & 84.8                                                        \\
M3DA            & \cmark & \xmark & \xmark & 72.8                                                                 & 98.4                                                                 & 81.3                                                                 & 96.1                                                                 & 89.6           & 87.6                                                       \\
MDDA             & \cmark & \xmark & \xmark & 78.6                                                                 & 98.8                                                                 & 79.3                                                                 & 93.9                                                                 & 89.7          & 88.1                                                        \\
LtC-MSDA         & \cmark & \xmark & \xmark & 85.6                                                                & 99                                                                   & 83.2                                                                 & 98.3                                                                 & 93             & 91.8                                                       \\
STEM         & \cmark & \xmark & \xmark & 89.7                                                                & 99.4                                                                   & 89.9                                                                 & 98.4                                                                 & 97.5             & 95.0                                                       \\\hline
SFDA             & \xmark & \cmark & \xmark & 86.2                                                                 & 95.4                                                                 & 57.4                                                                 & 95.8                                                                 & 84.8          & 83.9                                                        \\
SHOT             & \xmark & \cmark & \xmark & 90.4                                                                 & 98.9                                                                 & 58.3                                                                 & 97.7                                                                 & 83.9          & 85.8                                                        \\
MA               & \xmark & \cmark & \xmark & 90.8                                                                 & 98.4                                                                 & 59.1                                                                 & 98                                                                   & 84.5          & 86.2                                                        \\ \hline
DECISION         & \cmark & \cmark & \xmark & 93                                                                   & 99.2                                                                 & 82.6                                                                 & 97.8                                                                 & 97.5          & 94.0                                                        \\
CAiDA            & \cmark & \cmark & \xmark & 93.7                                                                 & 99.1                                                                 & 83.3                                                                 & \textbf{98.6}                                                        & \textbf{98.1} & 94.5                                                        \\ \hline
BASELINE & \multicolumn{3}{c}{(source-only)} & 53.2 & 97.6 & 63.5 & 90.5 & 87.1 & 78.5                                                       \\
\textbf{FREEDOM} & \cmark & \cmark & \cmark & \textbf{95.9}                                                        & \textbf{99.3}                                                        & \textbf{86.8}                                                        & 96.9                                                                 & 96.4          & \textbf{95.1}                                                        \\ \hline
\end{tabular}
\end{adjustbox}
\label{tab:fivedigit}
\end{table}

\begin{figure}[t!]
    \centering
    \includegraphics[width=\linewidth]{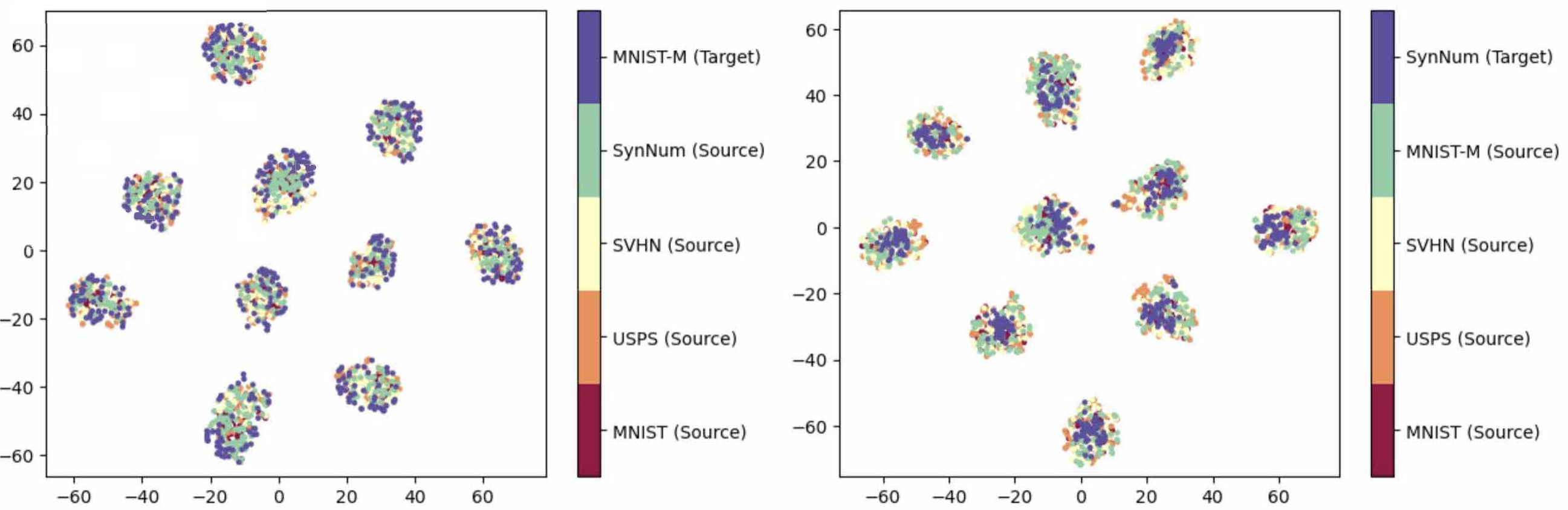}
        \caption{Class embedding space analysis with multiple sources and a target. (\textbf{Left}) Target is $\rightarrow$ \texttt{MM} (\textbf{Right}) Target is $\rightarrow$ \texttt{SYN}}
    \label{fig:class_tSNE}
    \vspace{-2mm}
\end{figure}

\subsubsection{Qualitative Analysis}

Besides the target adaptation performance, we need to glimpse if the FREEDOM model works well as we intended. 
To this end, we examine the target adaptation model's class embedding and style embedding spaces. 
FREEDOM aims to adapt a class encoder that transforms any target sample into the \textit{original space} discovered by source-side training.
In other words, we expect the class embedding space on the source and target sides to be identical. 
Figure \ref{fig:class_tSNE} demonstrates the tSNE plot of the class embedding of the source from its source-side model (baseline) and the target's class embedding from its adapted model; the plot results imply that the source and target class embeddings share the same space. 
Moreover, the class space is expected to have a different distribution for each class, and the result shows that the space not only the target and sources share them but also has ten independent class-conditional distributions.

\begin{figure}[t!]
    \centering
    \includegraphics[width=\linewidth]{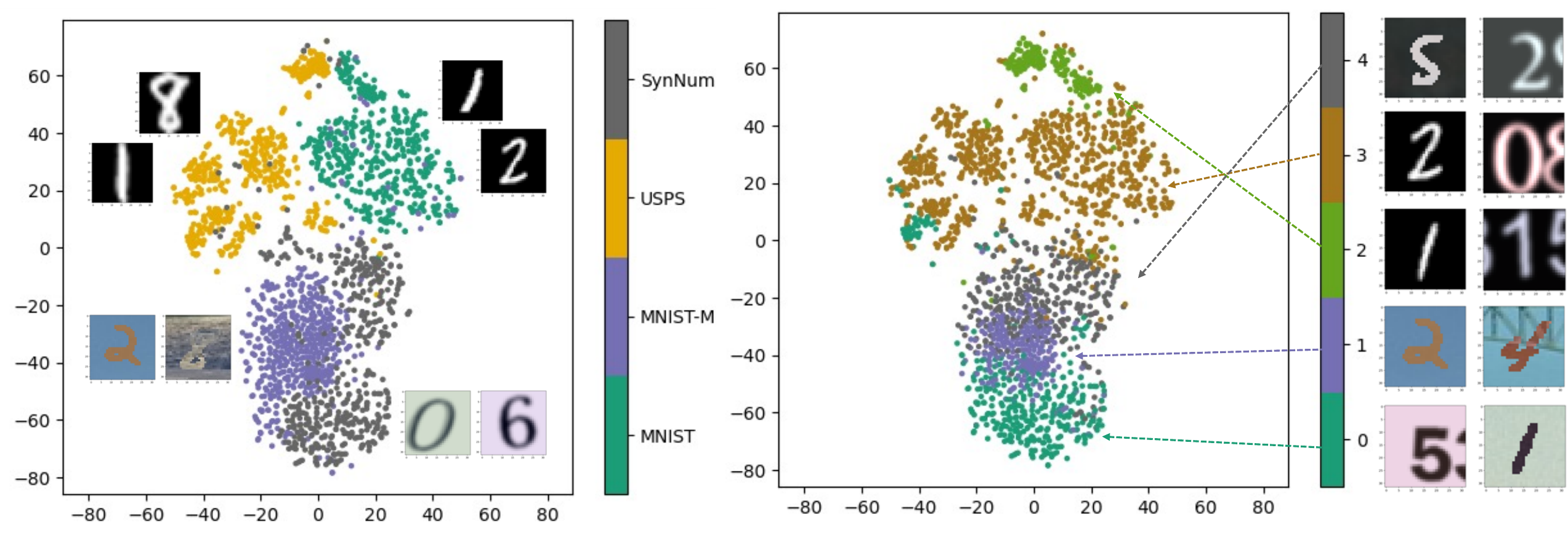}
    \caption{Style embedding space analysis ($\rightarrow$ \texttt{SV}) setting. (\textbf{Left}) tSNE plot with domain labels (\textbf{Right}) tSNE plot with style index inferenced by trained FREEDOM.}
    \label{fig:svhn_tSNE}
\end{figure}

\begin{figure}[t!]
    \centering
    \includegraphics[width=.94\linewidth]{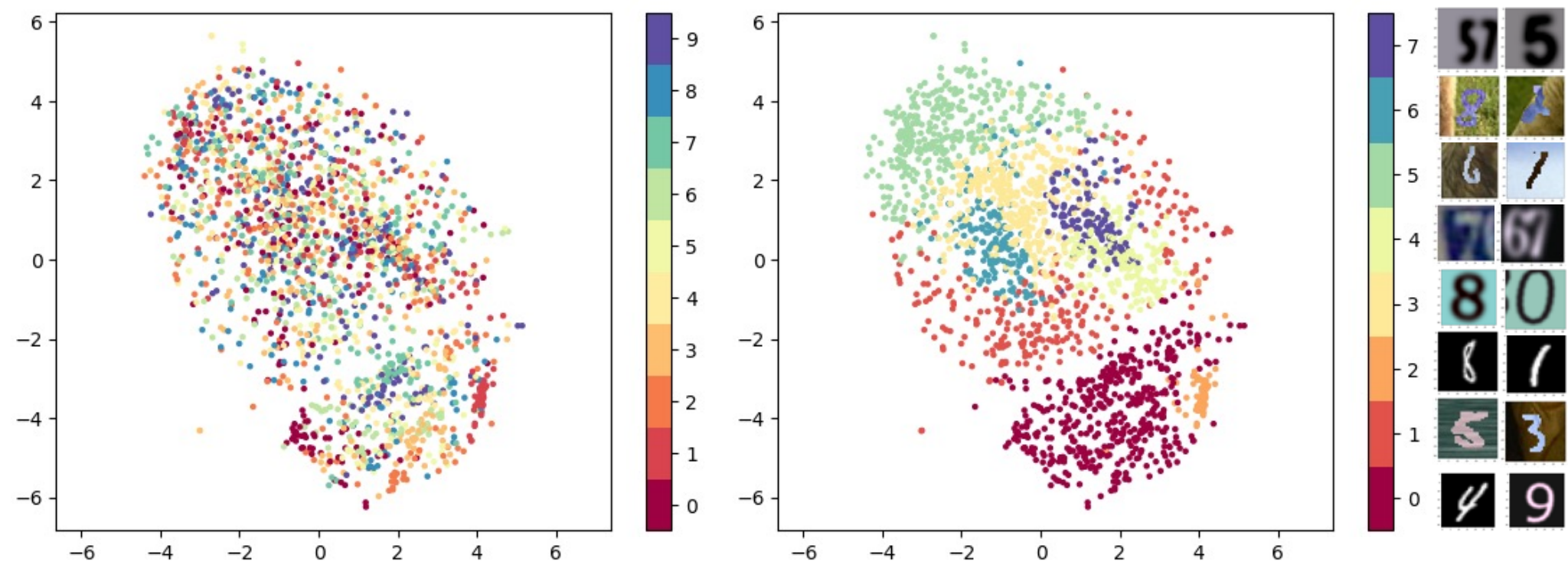}
    \caption{Style embedding space analysis with ($\rightarrow$ \texttt{UP}) setting. (\textbf{Left}) tSNE plot with class labels (\textbf{Right}) tSNE plot with style index inferenced by trained FREEDOM.}
    \label{fig:usps_tSNE}
    \vspace{3mm}
\end{figure}

We explore how the style encoder works. FREEDOM network disentangles data into style and class in order to match class space for both source and target by adapting the networks. Specifically, the style is defined as non-class knowledge completing the data distribution. In these senses, Figures \ref{fig:svhn_tSNE} and \ref{fig:usps_tSNE} show that FREEDOM's style encoder is trained as we intended.
The figures show the style embedding spaces induced by the source-trained style encoders for the cases where \texttt{SV} is the target and \texttt{UP} is the target, respectively. In both Figures, the embeddings look closer to domain-related information than class-related information. 

In Figure \ref{fig:svhn_tSNE} (left), one can see that the \texttt{SYN} domain data is divided into two groups by the \texttt{MM}, and the result in the figure on the right explains the reason for this. Despite the domain information set by humans, due to the various styles within one domain, FREEDOM recognized that there are a total of 5 styles, not four. Therefore, it identified the \texttt{SYN} domain as two different style groups, one with a dark background (style identifier: 4) and the other with a relatively light background color (style identifier: 0). 

Figure \ref{fig:usps_tSNE} is more intuitive to understand the \textit{non-class aspect} of style embedding. Both figures represent the same style of embedding space; only the legends are different. Fig. \ref{fig:usps_tSNE} (left) highlights the space with the class label as its legend, while the right shows the DPM model result from the style embedding, i.e., style identifier $s_n$. From this, we can confirm that the style embedding space can extract the rest of the information to restore the characteristics of the input image while being independent of the class as we intended.

\def\arraystretch{1.1}
\begin{table}[b!]
\centering
\caption{Experiment settings. }
\begin{adjustbox}{width=.85\linewidth}
\begin{tabular}{ccccc}
\hline
\multicolumn{5}{c}{\textbf{General setting}}                                                                                                                                                                               \\ 
               & \textbf{Learning Rate}                                                      & \textbf{Epoch}                                                           & \textbf{Warmup}                                                              & \textbf{batch size}       \\ \hline 
{\textbf{SRC}}   & 1e-3                                                               & 30                                                             & 10                                                                  & 256              \\
{\textbf{TGT}}   & 1e-3                                                               & 15                                                             & 5                                                                  & 256              \\ \hline 
\multicolumn{5}{c}{\textbf{Target configuration}}                                                                                                                                                                                   \\ 
               & \begin{tabular}[c]{@{}c@{}}$\alpha^{\text{class}}_{\text{recon}}$\end{tabular} & \begin{tabular}[c]{@{}c@{}}$\alpha^{\text{class}}_{\text{KL}}$\end{tabular} & \begin{tabular}[c]{@{}c@{}}$\alpha^{\text{class}}_{\text{helper}}$\end{tabular} & confidence level \\ \hline
{\textbf{conf1}} & 5                                                                  & 1                                                               & 10                                                                   & 0.3 
\\\hline
\end{tabular}
\end{adjustbox}
\label{tab:office-exp-conf}
\end{table}

\subsection{Evaluation on Office, Office-Caltech, and Office-Home}

\begin{table}[t!]
\caption{Evaluation results on \texttt{Office} dataset.}
\begin{adjustbox}{width=\columnwidth,center}
\begin{tabular}{cccccccc}
\hline
\textbf{Methods} & \textbf{MS}           & \textbf{SF}           & \textbf{DIF}          & \textbf{\begin{tabular}[c]{@{}c@{}}A,D\\ $\rightarrow$ W\end{tabular}} & \textbf{\begin{tabular}[c]{@{}c@{}}A,W\\ $\rightarrow$ D\end{tabular}} & \textbf{\begin{tabular}[c]{@{}c@{}}D,W\\ $\rightarrow$  A\end{tabular}} & \textbf{Avg.} \\ \hline
MDAN             & \cmark & \xmark & \xmark & 99.2                                                              & 95.4                                                                & 55.2                                                                & 83.2          \\
DCTN             & \cmark & \xmark & \xmark & 99.6                                                              & 96.9                                                                & 54.9                                                                & 83.8          \\
M3SDA            & \cmark & \xmark & \xmark & 99.4                                                              & 96.2                                                                & 55.4                                                                & 83.6          \\
MDDA             & \cmark & \xmark & \xmark & 99.2                                                              & 97.1                                                                & 56.2                                                                & 84.1          \\
LtC-MSDA         & \cmark & \xmark & \xmark & \textbf{99.6}                                                              & 97.2                                                                & 56.9                                                                & 84.5          \\ \hline
mDA              & \cmark & \xmark & \cmark & 93.1                                                              & 94.3                                                                & 64.2                                                                & 83.9          \\
MEC              & \cmark & \xmark & \cmark & 94.1                                                              & 95.1                                                                & 64.9                                                                & 84.7          \\ \hline
BAIT             & \xmark & \cmark & \xmark & 98.5                                                              & 98.8                                                                & 71.1                                                                & 89.4          \\
PrDA             & \xmark & \cmark & \xmark & 93.8                                                              & 96.7                                                                & 73.2                                                                & 87.9          \\
SHOT             & \xmark & \cmark & \xmark & 94.9                                                              & 97.8                                                                & 75                                                                  & 89.2          \\
MA               & \xmark & \cmark & \xmark & 96.1                                                              & 97.3                                                                & 75.2                                                                & 89.5          \\ \hline
DECISION         & \cmark & \cmark & \xmark & 98.4                                                              & 99.6                                                                & 75.4                                                                & 91.1          \\
CAiDA            & \cmark & \cmark & \xmark & 98.9                                                              & 99.8                                                                & 75.8                                                                & {91.5} \\ \hline
BASELINE & \multicolumn{3}{c}{(source-only)} & 97.8 & 99.0 & 76.8 & 91.2 \\
\textbf{FREEDOM} & \cmark & \cmark & \cmark & {98.1}                                                     & 99.0                                                      & \textbf{78.8}                                                               & \textbf{92.0} \\ \hline
\end{tabular}
\end{adjustbox}
\label{tab:office}
\end{table}

\begin{table}[t!]
\caption{Evaluation results on \texttt{Office-Caltech}.}
\begin{adjustbox}{width=\columnwidth,center}
\begin{tabular}{ccccccccc}
\hline
\textbf{Methods} & \textbf{MS}           & \textbf{SF}           & \textbf{DIF}          & \textbf{\begin{tabular}[c]{@{}c@{}}\textbf{A,D,C}\\ $\rightarrow$ W\end{tabular}} & \textbf{\begin{tabular}[c]{@{}c@{}} \textbf{A,C,W} \\ $\rightarrow$ D \end{tabular}} & \textbf{\begin{tabular}[c]{@{}c@{}} \textbf{C,D,W} \\ $\rightarrow$ A\end{tabular}} & \begin{tabular}[c]{@{}c@{}} \textbf{A,D,W} \\ $\rightarrow$ \textbf{C}\end{tabular} & \textbf{Avg.} \\ \hline
MDAN             & \cmark & \xmark & \xmark & 99.4                                                                & 98.7                                                                & 93.5                                                                & 91.6                                                       & 95.8        \\
DCTN             & \cmark & \xmark & \xmark & 99.4                                                                & 99.4                                                                & 94.1                                                                & 91.3                                                       & 96.05          \\
M3SDA            & \cmark & \xmark & \xmark & 99.5                                                                & 99.2                                                                & 94.5                                                                & 92.2                                                       & 96.35          \\
MDDA             & \cmark & \xmark & \xmark & 99.3                                                                & 99.6                                                                & 95.3                                                                & 92.3                                                       & 96.63        \\
LtC-MSDA         & \cmark & \xmark & \xmark & 99.4                                                                & 99.7                                                                & 93.7                                                                & 95.1                                                       & 96.98          \\ \hline
BAIT             & \xmark & \cmark & \xmark & 98.0                                                                & 97.5                                                                & 97.5                                                                & 95.7                                                       & 97.18        \\
PrDA             & \xmark & \cmark & \xmark & 97.6                                                                & 97.1                                                                & 97.3                                                                & 94.6                                                       & 96.65        \\
SHOT             & \xmark & \cmark & \xmark & 99.6                                                                & 96.8                                                                & 95.7                                                                & 95.8                                                       & 96.98          \\
MA               & \xmark & \cmark & \xmark & 99.8                                                                & 97.2                                                                & 95.7                                                                & 95.6                                                       & 97.08        \\ \hline
DECISION         & \cmark & \cmark & \xmark & 99.6                                                                & \textbf{100.0}                                                                & 95.9                                                                & 95.9                                                       & 97.85        \\
CAiDA            & \cmark & \cmark & \xmark & 99.8                                                                & \textbf{100.0}                                                                & 96.8                                                                & 97.1                                                       & \textbf{98.4}         \\ \hline
BASELINE & \multicolumn{3}{c}{(source-only)} & 98.3 & 98.4 & 95.1 & 92.0 & 96.0 \\
\textbf{FREEDOM} & \cmark & \cmark & \cmark & \textbf{100.0}                                                           & \textbf{100.0}                                                           & \textbf{97.1}                                                                     & 96.5                                                           & \textbf{98.4}     \\ \hline
\end{tabular}
\end{adjustbox}
\label{tab:office-caltech}
\end{table}

\begin{table}[t!]
\caption{Evaluation results on \texttt{Office-Home}.}
\begin{adjustbox}{width=\columnwidth,center}
\begin{tabular}{ccccccccc}
\hline
\textbf{Methods} & \textbf{MS}           & \textbf{SF}           & \textbf{DIF}          & \textbf{\begin{tabular}[c]{@{}c@{}}\textbf{A,C,P}\\ $\rightarrow$ R\end{tabular}} & \textbf{\begin{tabular}[c]{@{}c@{}} \textbf{A,C,R} \\ $\rightarrow$ P \end{tabular}} & \textbf{\begin{tabular}[c]{@{}c@{}} \textbf{A,P,R} \\ $\rightarrow$ C\end{tabular}} & \begin{tabular}[c]{@{}c@{}} \textbf{C,P,R} \\ $\rightarrow$ \textbf{A}\end{tabular} & \textbf{Avg.} \\ \hline
MDAN             & \cmark & \xmark & \xmark & 77.3                                                                & 77.6                                                                & 62.2                                                                & 65.4                                                       & 70.62        \\
DCTN             & \cmark & \xmark & \xmark & 78.7                                                                & 78.3                                                                & 63.8                                                                & 66.4                                                       & 71.8          \\
M3SDA            & \cmark & \xmark & \xmark & 79.4                                                                & 79.1                                                                & 63.5                                                                & 67.2                                                       & 72.3          \\
MDDA             & \cmark & \xmark & \xmark & 79.6                                                                & 79.5                                                                & 62.3                                                                & 66.7                                                       & 72.02        \\
LtC-MSDA         & \cmark & \xmark & \xmark & 80.1                                                                & 79.2                                                                & 64.1                                                                & 67.4                                                       & 72.7          \\ \hline
BAIT             & \xmark & \cmark & \xmark & 77.2                                                                & 79.4                                                                & 59.6                                                                & 71.1                                                       & 71.8        \\
PrDA             & \xmark & \cmark & \xmark & 76.8                                                                & 79.1                                                                & 57.5                                                                & 69.3                                                       & 70.7        \\
SHOT             & \xmark & \cmark & \xmark & 82.9                                                                & 82.8                                                                & 59.3                                                                & 72.2                                                       & 74.3          \\
MA               & \xmark & \cmark & \xmark & 81.7                                                                & 82.3                                                                & 57.4                                                                & 72.5                                                       & 73.5        \\ \hline
DECISION         & \cmark & \cmark & \xmark & 83.6                                                                & 84.4                                                                & 59.4                                                                & 74.5                                                       & 75.5        \\
CAiDA            & \cmark & \cmark & \xmark & 84.2                                                                & \textbf{84.7}                                                                & \textbf{60.5}                                                                & 75.2                                                       & \textbf{76.2}         \\ \hline
BASELINE & \multicolumn{3}{c}{(source-only)} & 82.5 & 82.8 & 51.3 & 71.6 & 72.3 \\
\textbf{FREEDOM} & \cmark & \cmark & \cmark & \textbf{84.6}                                                           & \textbf{84.7}                                                           & 56.0                                                                     & \textbf{75.7}                                                           & 75.3     \\ \hline
\end{tabular}
\end{adjustbox}
\label{tab:office-home}
\end{table}

We evaluate FREEDOM on \texttt{Office}, \texttt{Office-Caltech}, and \texttt{Office-Home} benchmarks with several random seeds; Tables \ref{tab:office}, \ref{tab:office-caltech} and \ref{tab:office-home} describe the results, respectively. \texttt{Office} and \texttt{Office-Caltech} datasets are similar in that they share three domains. We set the same configuration (\textbf{conf1} in Table \ref{tab:office-exp-conf}) for all targets.
Tables \ref{tab:office} and \ref{tab:office-caltech} show that the proposed FREEDOM outperforms existing source-free methods in an average value for the \texttt{Office} and the \texttt{Office-Caltech} datasets. Since these experiments take pre-trained ResNet as their backbone, their baseline already shows quite higher performance; for only shallow layers are adapted, there is not much room to improve further compared to the \texttt{Five-digit} benchmark. Interestingly, our baseline outperforms the competing methods even in some cases. It implies that the source-side learning algorithm proposed by FREEDOM finds a meaningful class embedding space, which makes adaptation stable.
On the other hand, the proposed method showed comparable accuracy in the \texttt{Office-Home} dataset. Even though its accuracy is not the utmost on average, it beats with three targets. Moreover, it can give comparable performance even though domain information is not provided.

\subsection{More analysis on FREEDOM}

In this section, we present additional empirical analysis to explore the potential of FREEDOM and its modules.

\subsubsection{Analysis of DPM-based Style Prior}
As an expedient to cope with domain information-free, we leveraged Dirichlet Process as style prior distribution of FREEDOM. To validate its efficacy, we conduct three experiments. First, we compare it with na\"ive Gaussian prior, which indicates a single multivariate Gaussian as a prior instead of DPM. Table \ref{tab:style_prior} demonstrates the comparison result. The results show that the DPM gives more margin for the style embedding and affects higher accuracy in unsupervised adaptation. 

\begin{figure}[t!]%
    \centering
    \subfloat[]{{\includegraphics[width=.49\linewidth]{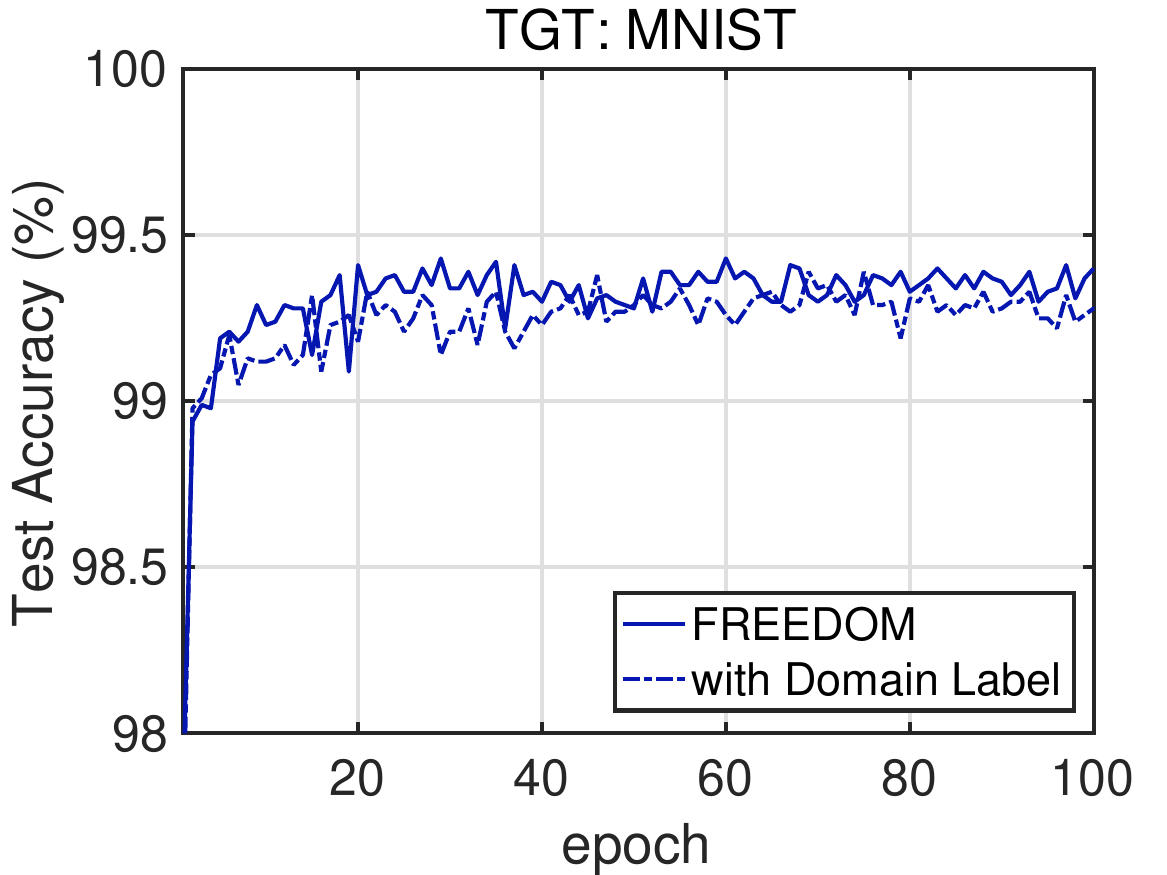} }} %
    \subfloat[]{{\includegraphics[width=.49\linewidth]{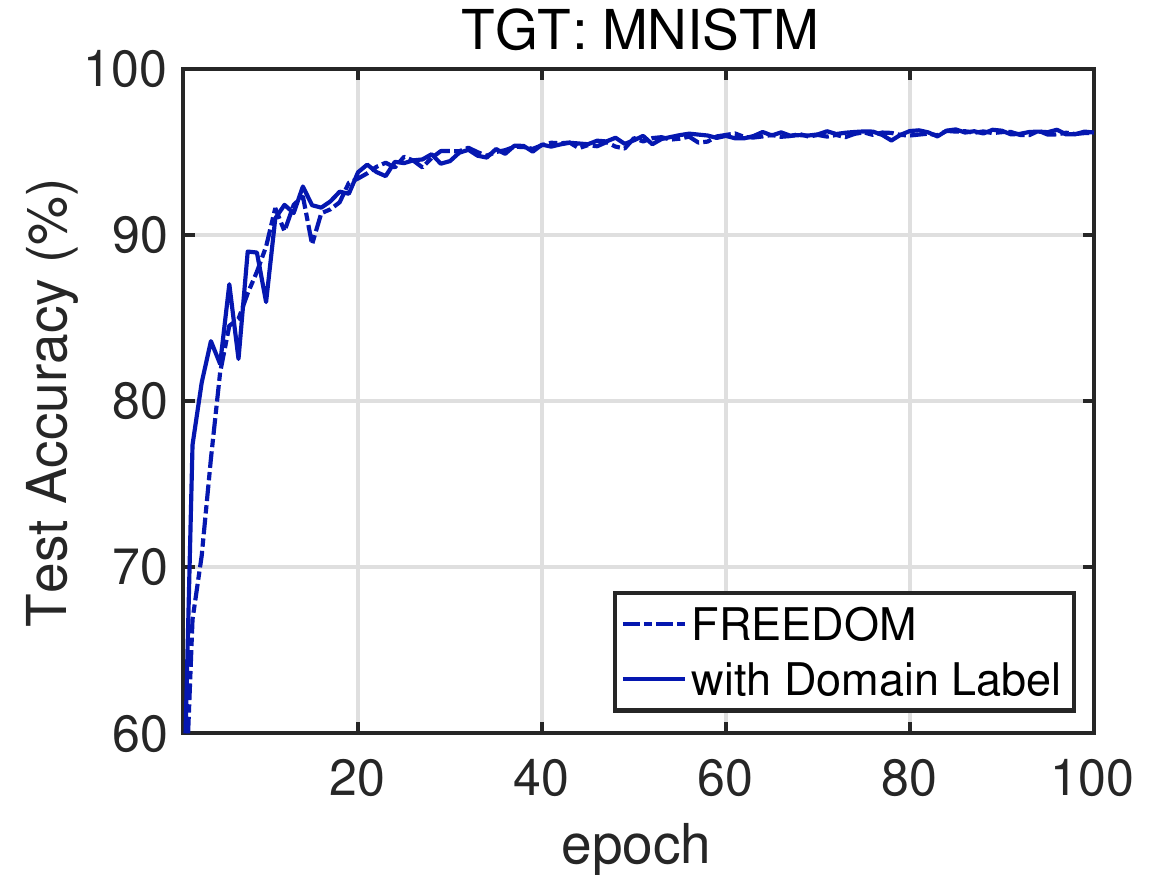} }}%
    \caption{Compare FREEDOM's target adaptation accuracy to the case using domain labels in style prior inference.}%
    \label{fig:style_domain_label_comparison}%
\end{figure}

We also compare the adaptation result with the case where domain information is given. We set different multivariate Gaussians for style embedding as we did for the class embedding. For example, we posit four different priors for style embedding for the \texttt{Five-digits} dataset. Fig. \ref{fig:style_domain_label_comparison} demonstrates the part of the result, demonstrating that FREEDOM is comparable to or even better than the case where the domain information is given. In the case where the target is SVHN, its test accuracy with DPM prior was improved by more than 1 \% poit.

Finally, we validate FREEDOM’s domain information-freeness by applying our method to the case where the source domain is configured with a single domain. We compare the performance with the prior work of SHOT\cite{shot-v119-liang20a}, of which the method is to serve the case of SFUDA. Table \ref{tab:single_source} demonstrates the result, that FREEDOM is suitable even for the single source case, free from the source domain information.

\subsubsection{Analysis of Final Model Size}
Existing MSFDA methods perform target adaptation by utilizing the ensemble of models learned from each domain model. Accordingly, the size of the target adaptation inference network increases with the number of source-domain. As shown in Table~\ref{tab:model_size}, existing techniques have different sizes of inference networks depending on the number of source-domain (3 for \texttt{Office} and 4 for \texttt{Office-Home}). Conversely, in FREEDOM, it can be seen that the size of the target adaptation inference network does not increase even if the number of source-domain increases; this gain becomes more prominent as the number of source domains increases. In addition, we want to emphasize that the final FREEDOM model is achievable without additional processing like knowledge distillation merely by discarding redundant parts of models, e.g., style encoder and decoder.

\begin{table}[t!]
\caption{Target accuracy comparison demonstrating efficacy of Bayesian non-parametric prior for style embedding.}
\begin{adjustbox}{width=.8\columnwidth,center}
\begin{tabular}{cccc}
\hline
\multicolumn{2}{c}{{{Five-digit (\%)}}}  & \multicolumn{2}{c}{{{Office (\%)}}}       \\ 
{Na\"ive Prior} & {DPM Prior} & {Na\"ive Prior} & {DPM} \\ \hline 
94.0                 & \textbf{95.6}               & 91.8                 & \textbf{92.7}               \\ \hline
\end{tabular}
\end{adjustbox}
\label{tab:style_prior}
\end{table}

\begin{table}[t!]
\caption{FREEDOM evaluation for the single source domain case}
\begin{adjustbox}{width=.8\columnwidth,center}
\begin{tabular}{cccccc}
\hline
  \textbf{Method}         & \textbf{A→W}           & \textbf{W→A}           & \textbf{D→A }          & \textbf{A→D}           & \textbf{Avg.}          \\ \hline\hline
SHOT-IM    & 91.2          & 71.4          & 72.5          & 90.6          & 81.4          \\
SHOT(full) & 90.1          & 74.3          & \textbf{74.7} & 94.0          & 83.3          \\ \hline
FREEDOM    & \textbf{91.3} & \textbf{75.2} & 71.3          & \textbf{96.0} & \textbf{83.5} \\ \hline
\end{tabular}
\end{adjustbox}
\label{tab:single_source}
\end{table}

\def\arraystretch{1.2}
\begin{table}[t!]
\caption{Inference network size (MB) comparison.}
\begin{adjustbox}{width=.8\columnwidth,center}
\begin{tabular}{cccc}
\hline
\textbf{SCENARIO} & {\textbf{METHOD}}                                                     & \texttt{\textbf{Office}}       & \texttt{\textbf{Office-Home}}  \\ \hline \hline
\multicolumn{2}{c}{\textbf{Number of Source Domains}}                                      & 3        & 4        \\ \hline
SFUDA                  & SHOT                                                     & 195.4        & 293.1        \\ \hline
\multirow{2}{*}{MSFDA} & DECISION                                                 & 191.4        & 287.1        \\ 
                       & CAiDA                                                    & \textbf{183.8 }            &    275.8          \\ \hline
TFDA               & FREEDOM & {219.4} & \textbf{219.4} \\ \hline
\end{tabular}
\end{adjustbox}
\label{tab:model_size}
\end{table}

\begin{figure}[t!]
    \centering
    \includegraphics[width=.7\linewidth]{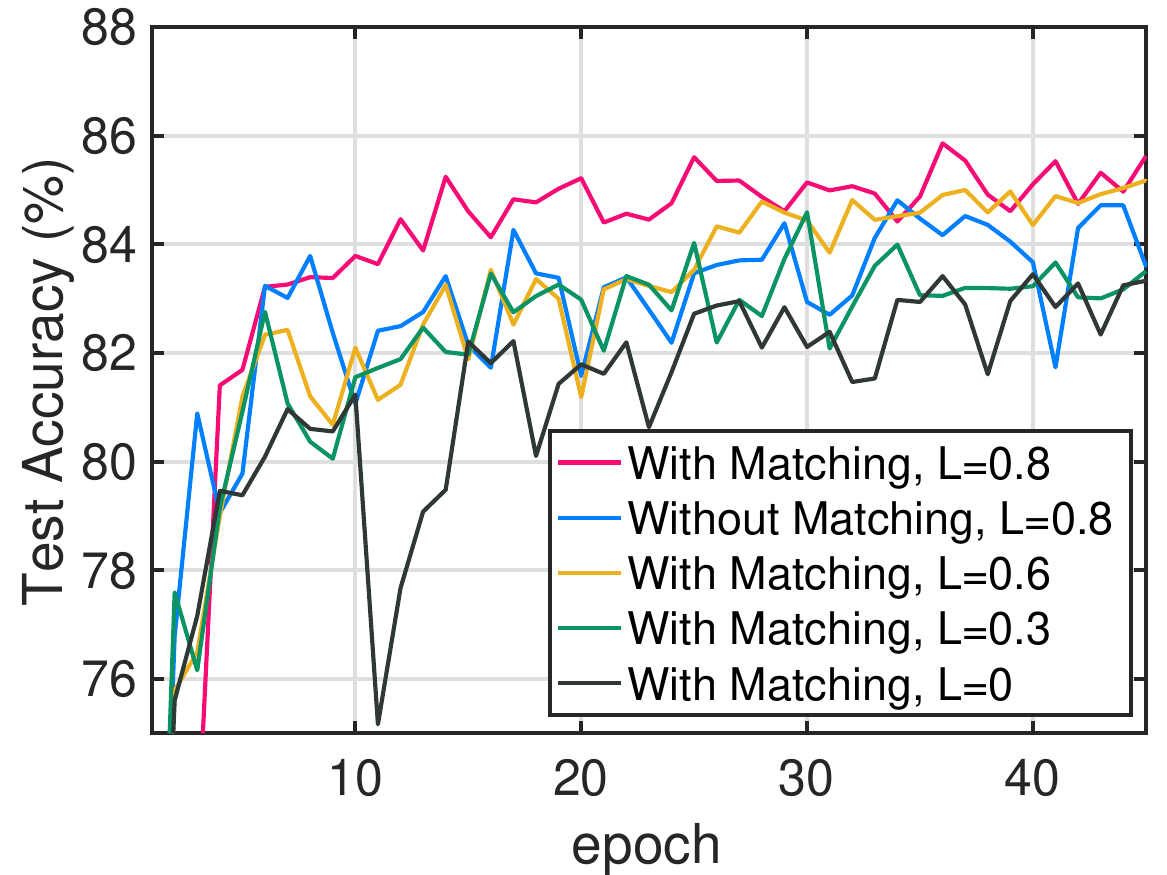}
    \caption{Effect of confident data selection. Comparison for 1) using prediction matching, i.e., $\arg\max \gamma^* = \arg\max \hat{\bm{y}}$ 2) different confidence level $L = \{0.8, 0.6, 0.3, 0\}$, tested with ($\rightarrow$ \texttt{SVHN}) case.}
    \label{fig:confidence_batch_selection}
\end{figure}

\subsubsection{Analysis of the effect of batch selection}
For the stable target adaptation, we introduce two batch selection strategies in FREEDOM: 1) filtering out batches using agreement tests on moment-based inference and classifier likelihood inference and 2) confidence-based filtering.
Figure \ref{fig:confidence_batch_selection} contains with and without matching cases with the same confidence level. It demonstrates that the matching-based selection provides a more stable adaptation than without it.
In addition, we compare five different settings on confident batch selection to validate the rationale for using label smoothing loss in source-side training and data selection leveraging on it. According to \cite{muller2019does}, label smoothing loss can calibrate the network, where its prediction softmax denotes confidence in the inference. Thus, using the calibrated network, FREEDOM filters out target data samples to be used in pseudo-label inference. To validate its efficacy, we set four different confidence levels $\{0.8, 0.6, 0.3, 0\}$ to compare their adaptation.
Figure \ref{fig:confidence_batch_selection} demonstrates that the target selection based on a higher confidence level (0.8 or 0.6) provides a more stable adaptation than the lower one (0.3 or 0).

\subsubsection{Abalation Study}
We analyze each FREEDOM module's efficacy through an ablation study. In target adaptation, FREEDOM introduces several submodules, utilizing its generative model. We measure the performance change in the absence of each module in the proposed overall target adaptation algorithm (Algorithm \ref{algo:target-side}), as shown in Table \ref{tab:ablation}. 
The consupicuous performance degradation in \textbf{a) warm-up}  highlights the necessity of the alternating adaptation algorithm in target adaptation(\texttt{line 2} in Algorithm \ref{algo:target-side}).
In other words, adapting the class encoder to the target without sequential optimization of the target data shows that the maximum likelihood loss can hinder regular training. 

The results of \textbf{b) without class-prototype learning} validates the class regularization term in Eq. (\ref{eq:tgt_class}). 
One of FREEDOM's main strategies is transferring the class-conditional distribution learned from the source side to the target. 
We specifically add an $\mathcal{L}_{\text{KL}}^{\text{class}}$ regularization term for this strategy when adapting the target's class encoder. Excluding this term may cause 6.9\% point degradation of the final accuracy.

Finally, \textbf{c) without confidence level} and \textbf{d) without matching }show the effect of batch selection. As explained in the previous section, batch selection determines how good pseudo-label data can be provided when FREEDOM performs target adaptation according to data characteristics. Confidence level-based and matching-based filtering both utilize FREEDOM's generative model characteristics, showing that learning performance can be further improved when both are used.

\section{Conclusion}
In this paper, we first propose a more pragmatic scenario named TFDA, which relaxes the two significant obstacles, information of 1) domain label and 2) the number of domains, for applying domain adaptation to AI-based services.
This relaxation reduces the amount of information necessary for training, thus introducing more practicality. On the other hand, this relaxation enforces the network to learn without domain labels, which is a non-trivial problem to solve. Our proposed method, FREEDOM, resolve the hurdles by disentangling the class features and style features and applying bayesian non-parametric modeling on the style features. We evaluate FREEDOM on four popular MSDA benchmarks to validate our method. We further demonstrate the feasibility of each module of the proposed technique through experiments on embedding space and various ablation studies.

\begin{table}[t!]
\centering
\caption{Ablation study results on SVHN target}
\begin{tabular}{clc}
\hline
\multicolumn{2}{c}{\textbf{Method}}                & \textbf{→ SV} \\ \hline \hline
DECISION                 &                         & 82.6          \\
CAiDA                    &                         & 83.3          \\ \hline
\multirow{5}{*}{FREEDOM} & a) w/o warm-up          & 19.6          \\
                         & b) w/o class-prototype  & 79.9          \\
                         & c) w/o confidence level & 83.1          \\
                         & d) w/o matching         & 84.6          \\
                         & e) with all (proposed)  & \textbf{86.8}          \\ \hline
\end{tabular}
\label{tab:ablation}
\end{table}

\appendices
\section{Proof of Lemma}

\textbf{Lemma 1.} The optimal variational posterior of the style identifier $s$ is given as 
\begin{align*}
    q^*(s\vert\bm{x}) = \mathbb{E}_{q_{\bm{\Phi}^{\text{style}}}(\bm{z}^{\text{style}}\vert\bm{x})}[p(s\vert\bm{z}^{\text{style}})].
\end{align*}

\textit{Proof of Lemma 1.}
\begin{align*}
    & \mathcal{L}_{\text{ELBO}}^{\text{SRC}}(\bm{x}) = \mathbb{E}_{q(\bm{z}^{\text{style}},\bm{z}^{\text{class}},s,y\vert\bm{x})}\Big[\log \frac{p(\bm{x},\bm{z}^{\text{style}}, \bm{z}^{\text{class}},s,y)}{q(\bm{z}^{\text{style}},\bm{z}^{\text{class}},s,y\vert\bm{x})}\Big] \\
    & =\mathbb{E}_{q(\cdot\vert\bm{x})}\Big[\log \frac{p(\bm{x} \vert \bm{z}^{\text{style}}, \bm{z}^{\text{class}})p(\bm{z}^{\text{class}}\vert y) p(y) p(\bm{z}^{\text{style}}\vert s) p(s)}{q(\bm{z}^{\text{style}},\bm{z}^{\text{class}},s,y\vert \bm{x})}\Big] \\
    & =\mathbb{E}_{q(\cdot\vert\bm{x})}\Big[\log \frac{p(\bm{x} \vert \bm{z}^{\text{style}}, \bm{z}^{\text{class}})p(\bm{z}^{\text{class}}\vert y) p(y) p(s \vert \bm{z}^{\text{style}}) p(\bm{z}^{\text{style}})}{q(\bm{z}^{\text{style}},\bm{z}^{\text{class}},s,y\vert \bm{x})}\Big] \\
    & = \underbrace{\mathbb{E}_{q(\cdot\vert\bm{x})} \Big[ \log \frac{p(\bm{x},\bm{z}^{\text{style}},\bm{z}^{\text{class}}, y)}{q(\bm{z}^{\text{style}}, \bm{z}^{\text{class}}, y\vert\bm{x})}\Big]}_{\circ} - \mathbb{E}_{q(\cdot\vert\bm{x})}\Big[\log\frac{q(s\vert\bm{x})}{p(s\vert\bm{z}^{\text{style}})}\Big]\\
    & = \circ  - \int q(\bm{z}^{\text{style}}\vert\bm{x}) \sum_s q(s\vert\bm{x}) \Big[\log\frac{q(s\vert\bm{x})}{p(s\vert\bm{z}^{\text{style}})}\Big] d\bm{z}^{\text{style}}\\
    & = \circ  - \mathbb{E}_{q(\bm{z}^{\text{class}}\vert\bm{x})}[\mathcal{D}_{\text{KL}}(q(s\vert\bm{x}) \vert\vert p(s\vert\bm{z}^{\text{style}}))],
\end{align*}
where $\circ$ is the term extraneous to the style identifier $s$. From the above, one can find that the optimal variational posterior of $s$ can be obtained when $\mathcal{D}_{\text{KL}}(q(s\vert\bm{x}) \vert\vert p(s\vert\bm{z}^{\text{style}})) = 0$, for the KL divergence is always non-negative. In addition, we know that $\sum_s q(s\vert\bm{x}) = \sum_s p(s\vert\bm{z}^{\text{style}}) = 1$, leading to $q(s\vert\bm{x}) = p(s\vert\bm{z}^{\text{style}})$. We draw the conclusion by taking expectation on both side, i.e., $q^*(s\vert\bm{x}) = \mathbb{E}_{q(\bm{z}^{\text{style}}\vert\bm{x})}[p(s\vert\bm{z}^{\text{style}})]$ $\blacksquare$

\section{Details of Loss derivation }

\subsubsection{Class regularization loss}
\label{app:class_regularization_loss}
 The following Lemma 2 supports to the derivation of the tractable form of the class regularization loss.

\textbf{Lemma 2.} For the given two multi-variate Gaussian, $p(\bm{x}) = \mathcal{N}(\bm{x};\bm{\mu}, \bm{\Sigma})$ and $q(\bm{x}) = \mathcal{N}(\bm{x};\bm{\hat{\mu}},\bm{\hat{\Sigma}})$,
\begin{align*}
    & \mathbb{E}_{q(\bm{x})}[\log p(\bm{x})]= \\
    & -\frac{1}{2} \sum_{h=1}^{H} \Big(\log 2\pi\bm{\Sigma}\vert_h + \frac{\bm{\hat{\Sigma}}\vert_h}{\bm{\Sigma}\vert_h} +\frac{(\hat{\bm{\mu}}\vert_h-\bm{\mu}\vert_h)^2}{\bm{\Sigma}\vert_h} \Big),
\end{align*}
where $H$ is the dimension of $\bm{x}$, $\bm{\Sigma}\vert_h$ is the $(h,h)^{\text{th}}$ element of the diagonal matrix $\bm{\Sigma}$, and $\bm{\mu}\vert_h$ denote the $h^{\text{th}}$ element of vector $\bm{\mu}$.

Then, one can obtain the tractable form of the class regularization loss.
\begin{equation}
\begin{aligned}
& \mathcal{L}_{\text{KL}}^{\text{class}}(\bm{x}, y) =
\frac{1}{2} \sum_{h=1}^{H_c} \Big(
\log (2\pi \bm{\Sigma}_y^{\text{class}}\vert_h)
+\frac{\hat{\bm{\Sigma}}^{\text{class}}\vert_h}{\bm{\Sigma}_y^{\text{class}}\vert_h} \\& + \frac{(\bm{\hat{\mu}}^{\text{class}}\vert_h + \bm{\mu}_y^{\text{class}}\vert_h)^2}{\bm{\Sigma}_y^{\text{class}}\vert_h} 
 - \log 2\pi \bm{\hat{\Sigma}}^{\text{class}}\vert_h - 1
\Big) - \log \pi_y^{\text{class}}
\end{aligned}
\end{equation}

\subsubsection{Style regularization loss}
The style regularization loss is computed with the given prior parameters obtained by the variational inference on the DPM; its tractable form is derived with Lemma 2 as well. Here is the loss function:
\begin{equation}
\begin{aligned}
    &\bar{\mathcal{L}}_{\text{KL}}^{\text{Style}}(\bm{x}, \bm{\beta}^*, \bm{\mu}^*, \bm{\Sigma}^*) \\
    & = \mathbb{E}_{q}[\log q(\bm{z}^{\text{style}}\vert \bm{x})] - \mathbb{E}_{q}[\log p(\bm{z}^{\text{style}}\vert \bm{\mu}^*, \bm{\Sigma}^*)] \\
    & = -\frac{H_s}{2} - \frac{1}{2}\sum_{h=1}^{H_s} \Big( \log 2\pi \hat{\Sigma}\vert_h + \log 2\pi \Sigma_{s_n^*}\vert_h + \frac{\hat{\Sigma}\vert_h}{\Sigma_{s_n^*}\vert_h} \\
    & \indent + \frac{(\hat{\mu}\vert_h - \mu_{s_n^*}\vert_h)^2}{\Sigma_{s_n^*\vert_h}}\Big).
\end{aligned}
\end{equation}

\subsubsection{Class label inference with original distribution}
\label{app:gamma}
From the generative model of class embedding, one can infer the most probable class label. The inference on the class label from an input can be drawn similarly to Lemma 1. If we modify the ELBO in terms of the class label $y_n$, we can derive the optimal inference as $q^*(y_n\vert\bm{x}_n) = \mathbb{E}_{q(\bm{z}_n^{\text{class}}\vert\bm{x}_n)}[p(y_n\vert\bm{z}_n^{\text{class}})]$, which can be further approximated with the reparameterization trick and \cite{jiang2017variational} as follows
\begin{align*}
   \mathbb{E}_{q(\bm{z}^{\text{class}}\vert \bm{x})}[p(y\vert\bm{x})] \approx \frac{1}{L} \sum_{l=1}^{L} \Big[\frac{p(y_k)p(\bm{z}^{{\text{class}}^{(l)}}\vert y_k)}{\sum_{j=1}^C p(y_j) p(\bm{z}^{{\text{class}}^{(l)}}\vert y_j)} \Big]_{k=1}^{C},
\end{align*}
where $\bm{z}^{{\text{class}}^{(l)}} = \bm{\mu}^{\text{class}} + \bm{\Sigma}^{\text{class}} \circ \epsilon ^{(l)}$ and $\epsilon ^{(l)}$ is the random noise following the normal distribution.

\bibliographystyle{IEEEtran}
\bibliography{./reference.bib}

\end{document}